\let\mypdfximage\pdfximage
\def\pdfximage{\immediate\mypdfximage}

\documentclass[journal,onecolumn,10pt]{IEEEtran1}

\makeatletter
\def\markboth#1#2{\def\leftmark{\@IEEEcompsoconly{\sffamily}\MakeUppercase{\protect#1}}%
\def\rightmark{\@IEEEcompsoconly{\sffamily}\MakeUppercase{\protect#2}}}
\makeatother

\let\labelindent\relax

\usepackage{times}
\usepackage{marvosym}
\usepackage{longtable}
\usepackage{graphics}
\usepackage{graphicx}
\usepackage{caption}
\usepackage[english]{babel}
\usepackage{subfig}
\usepackage{epsfig}
\usepackage{color}
\usepackage{multirow}
\usepackage{mathrsfs}
\usepackage{textcomp}
\usepackage{amsfonts}
\usepackage{amsmath}
\usepackage{amssymb}
\usepackage{nccmath}
\usepackage[numbers,square,sort&compress,comma]{natbib}
\usepackage{chapterbib}
\usepackage[ruled,vlined,linesnumbered]{algorithm2e}
\usepackage{setspace}
\usepackage{verbatim}
\usepackage{wrapfig}
\usepackage{dblfloatfix}
\usepackage{comment}
\usepackage[strict]{changepage}
\usepackage{ragged2e}
\usepackage{microtype}
\usepackage{enumitem}
\usepackage{nccmath}
\usepackage{hyperref}
\usepackage{doi}
\usepackage{afterpage}
\usepackage{multirow,bigdelim}
\usepackage{booktabs}
\usepackage{color}
\usepackage[outline]{contour}
\usepackage{xcolor}
\usepackage{bibunits}
\usepackage[nameinlink]{cleveref}

\setlist{parsep=0pt,listparindent=\parindent}

{\fontsize{9.75}{10}\selectfont \defaultbibliographystyle{Sledge-TPAMI-2021-1col} \defaultbibliographystyle{IEEEtran}}

\DeclareCaptionFont{singlespacing}{\setstretch{1}}
\numberwithin{figure}{section}
\renewcommand{\thefigure}{\arabic{section}.\arabic{figure}}

\renewcommand \thesection{\arabic{section}}
\numberwithin{equation}{section}

\hypersetup{bookmarks=false,bookmarksopen=false,pdfpagemode=empty,pdfstartview=}

\let\oldnl\nl
\newcommand{\nonl}{\renewcommand{\nl}{\let\nl\oldnl}}

\title{\singlespacing\sf\huge External-Memory Networks for Low-Shot Learning of Targets in Forward-Looking-Sonar Imagery}
\author{Isaac J. Sledge, \emph{Member, IEEE}, Christopher D. Toole, \emph{Member, IEEE},\\ Joseph A. Maestri, \emph{Member, IEEE}, and Jos\'{e} C. Pr\'{i}ncipe, \emph{Life Fellow, IEEE}%
\thanks{\fontdimen2\font=1.55pt Isaac J. Sledge is the Senior Machine Learning Scientist and Dr. Delores M. Etter Assistant Secretary of the Navy Emergent Engineer with the Advanced Signal Processing and Automated Target Recognition Branch, Naval Surface Warfare Center, Panama City, FL, USA (email: isaac.j.sledge@navy.mil).  He the director of the Machine Intelligence Defense (MIND) lab at the Naval Sea Systems Command}
\thanks{\fontdimen2\font=1.55pt Christopher D. Toole is a Scientist with the Signal Processing and Analysis Branch, Naval Undersea Warfare Center, Newport, RI, USA (email: christopher.d.toole1@navy.mil).}
\thanks{\fontdimen2\font=1.55pt Joseph A. Maestri is an Engineer with the Signal Processing and Analysis Branch, Naval Undersea Warfare Center, Newport, RI, USA (email: joseph.maestri@navy.mil)}
\thanks{\fontdimen2\font=1.55pt Jos\'{e} C. Pr\'{i}ncipe is the Don D. and Ruth S. Eckis Chair and the Distinguished Professor with both the Department of Electrical and Computer Engineering and the Department of Biomedical Engineering, University of Florida, Gainesville, FL, USA (email: principe@ufl.edu).  He is the director of the Computational NeuroEngineering Laboratory (CNEL) at the University of Florida.\vspace{0.1cm}}
\thanks{The work of the first and fourth authors was funded by grants N00014-19-WX-00636 (Marc Steinberg), N00014-21-WX-00476 (J. Tory Cobb), N00014-21-WX-00525 (Thomas McKenna), and N00014-21-WX-01348 (Marc Steinberg) from the US Office of Naval Research.  The first author was also supported by in-house laboratory independent research (ILIR) grant N00014-19-WX-00687 (Frank Crosby) from the US Office of Naval Research and a Naval Innovation in Science and Engineering (NISE) grant from NAVSEA.  The work of the second and third authors was supported by a Naval Innovation in Science and Engineering (NISE) grant from NAVSEA.}%
}
\begin{document}
\begin{bibunit}
\bstctlcite{IEEEexample:BSTcontrol}

\maketitle
\RaggedRight\parindent=1.5em
\fontdimen2\font=2.1pt
\vspace{-1.55cm}\begin{abstract}\normalsize\singlespacing
\vspace{-0.25cm}{\small{\sf{\textbf{Abstract}}}}---We propose a memory-based framework for real-time, data-efficient target analysis in forward-looking-sonar (FLS) imagery.  Our framework relies on first removing non-discriminative details from the imagery using a small-scale {\sc DenseNet}-inspired network.  Doing so simplifies ensuing analyses and permits generalizing from few labeled examples.  We then cascade the filtered imagery into a novel {\sc NeuralRAM}-based convolutional matching network, {\sc NRMN}, for low-shot target recognition.  We employ a small-scale {\sc FlowNet}, {\sc LFN} to align and register FLS imagery across local temporal scales.  {\sc LFN} enables target label consensus voting across images and generally improves target detection and recognition rates.

We evaluate our framework using real-world FLS imagery with multiple broad target classes that have high intra-class variability and rich sub-class structure.  We show that few-shot learning, with anywhere from ten to thirty class-specific exemplars, performs similarly to supervised deep networks trained on hundreds of samples per class.  Effective zero-shot learning is also possible.  High performance is realized from the inductive-transfer properties of {\sc NRMN}s when distractor elements are removed.
\end{abstract}%
\begin{IEEEkeywords}\normalsize\singlespacing
\vspace{-1.35cm}{{\small{\sf{\textbf{Index Terms}}}}---Memory network, meta-learning, learning-to-learn, lifelong learning, target detection, target recognition, forward-looking sonar, imaging sonar}
\end{IEEEkeywords}
\IEEEpeerreviewmaketitle
\allowdisplaybreaks
\singlespacing

\vspace{-0.4cm}\subsection*{\small{\sf{\textbf{1.$\;\;\;$Introduction}}}}\addtocounter{section}{1}

Analyzing properties of underwater targets is important for a variety of applications \cite{SledgeIJ-jour2020a}.  Traditional approaches to these tasks often leverage side-scan, circular-scan, and spiral-scan sonar imagery.  While such modalities are suitable for stationary targets, they cannot handle mobile targets well.

Forward-looking sonar (FLS) imagery is well suited for stationary and mobile targets \cite{CobbJT-conf2005a,PetillotY-jour2001a,QuiduI-jour2012a}, let alone mapping \cite{FranchiM-conf2018a,FranchiM-conf2020a,HurtosN-conf2013a,FranchiM-conf2019a,HensonBT-jour2019a}.  A variety of deep and non-deep approaches have achieved remarkable target detection, tracking, and recognition rates for this modality (see section 2).  Those belonging to the former category are largely preferred, currently, over the latter.  However, such approaches are largely supervised and hence commonly require large amounts of labeled data to perform well.  This property severely limits their scalability, even when pre-training.  It also limits their applicability to novel and rare target categories where FLS imagery may either be scarce or even not exist.

To overcome such issues, we believe that both zero- and low-shot learning strategies should be adopted within deep networks.  Doing so would permit low-sample generalization with learned feature representations.  It would also sidestep dense annotation requirements.

Low-shot learning \cite{SunQ-conf2019a,JamalMA-conf2019a,LiuB-conf2020a,YuZ-conf2020a,ElskenT-conf2020a} seeks to distinguish between novel classes from few labeled samples, which is difficult for most deep networks to do well \cite{EdwardsH-conf2017a}.  Conventional low-shot strategies often partition the training process into two phases.  The first of these is an auxiliary meta-learning stage \cite{MunkhdalaiT-conf2017a} wherein transferrable knowledge representations are discerned \cite{VinyalsO-coll2016a,SnellJ-coll2017a}.  The second is a learning and fine-tuning process \cite{FinnC-conf2017a}, which entails either updating network parameters \cite{DouzeM-conf2018a,QiH-conf2018a}, via gradient-based optimization \cite{RaviS-conf2017a}, or holding them fixed \cite{BertinettoL-coll2016a} and updating some stored characterization \cite{SantoroA-conf2016a,KaiserL-conf2017a}.  Zero-shot learning \cite{SocherR-coll2013a,WanZ-coll2019a} is an extreme instance of low-shot learning.  Only a single example is provided when conducting zero-shot learning \cite{XianY-conf2017a,KodirovE-conf2017a,XianY-conf2018a}, which is typically in the form of a text description that relates a new class to one that was previously observed during training \cite{FarhadiA-conf2009a,HwangSJ-conf2011a,AkataZ-conf2013a}.  Note that data augmentation \cite{CubukED-conf2019a}, as it is commonly implemented, cannot be readily applied to address zero- and low-shot learning; generated data \cite{WangYX-conf2018a}, let alone unlabeled data \cite{RenM-conf2018a,GarciaV-conf2018a}, would be more appropriate for this task, even though they also have issues \cite{ZhangJ-conf2019a}.

Progress has been made in addressing low-sample-learning problems for deep networks.  Most existing strategies work well only for single-target samples where targets are well localized, though.  It is doubtful that they could scale to forward-looking sonar imagery.  Foremost, the imagery, with respect to the anticipated target size, is quite large.  Secondly, one image can contain multiple target types.

Here, we propose an end-to-end-trainable, memory-augmented framework that conducts real-time, low-shot learning on FLS images (see section 3).  This framework and many of its components are novel.  It is also the first instance of low-shot learning for any imaging sonar modality.

In the first part of this framework, we use a pre-trained, small-scale {\sc DenseNet}-inspired \cite{HuangG-conf2017a} network, {\sc DNSS}, for saliency-based segmentation of target-like objects.  {\sc DNSS} significantly simplifies feature learning, as distractor components from the FLS imagery are largely removed.  Detected target regions are then presented to a matching network, {\sc NRMN}, which provides multi-target, multi-class recognition.  {\sc NRMN} learns FLS image features that, for the training set, are used by a memory controller to populate a {\sc NeuralRAM} \cite{WestonJ-conf2015a,GuicehreC-jour2018a} recurrent memory.  This memory-stored representations can be used for matching-based recognition of multiple target classes.  A convolutional-based similarity measure is learned to map between the training set and support set representations to discern the appropriate class.  If no good matches are found in memory, then the {\sc NRMN}s will return that an unknown class has been discovered \cite{BendaleA-conf2016a,WangY-conf2018a,YoshihashiR-conf2019a}, which allows for sparse, human-in-the-loop annotation.  To handle temporal label propagation across consecutive FLS images, we leverage a pre-trained, small-scale \cite{HuiTW-conf2018a} {\sc FlowNet} \cite{DosovitskiyA-conf2015a,IlgE-conf2017a}, which we call {\sc LFN}.  This network helps retain contextual target focus over time, which enables tracking mobile targets for a mostly stationary sensor platform and mostly stationary targets for a mobile sensor platform.

Our {\sc NRMN}s are naturally related to networks that aim to uncover an effective representation for low-shot comparisons.  Many of these works learn a transferrable embedding \cite{KochG-conf2015a,VinyalsO-coll2016a} and utilize a pre-defined metric for matching \cite{SnellJ-coll2017a}.  We, however, take the view that both a deep embedding \cite{WangX-conf2019a} and an associated deep, non-parametric similarity measure \cite{SungF-conf2018a,HaoF-conf2019a} for relating embedded representations should be discerned.  This measure can be either between sample pairs or between samples and descriptions, which, respectively, facilitates low-shot and zero-shot learning.  Regardless of the choice, by adopting a deep, data-driven solution, we sufficiently enhance the {\sc NRMN}s' inductive bias so that generalizable target classifiers can be easily learned.  Although measures between point-estimated representations may be vulnerable to noise \cite{ZhangJ-conf2019a}, a concern in sample-scarce settings, we find it does not adversely impact our results.  In fact, learning a measure for querying the neural memories noticeably helps compared to the fixed-metric case \cite{CaiQ-conf2018a}.  It works in tandem with the feature-embedding process to improve discriminability.  Fixed metrics, which are commonly used in memory networks, pre-suppose near-linear separability of the features.  This property is rarely satisfied in practice.

We empirically assess our framework on real-world FLS imagery (see section 4).  We illustrate that our {\sc DNSS} networks can isolate targets well, compared to alternate supervised and unsupervised schemes, even in the presence of distractors.  This pre-processing step allows the {\sc NRMN}s to achieve detection and recognition rates that are on par with conventional deep networks trained on many times more samples.  The performance of {\sc NRMN}s compares favorably to alternate low-shot-learning strategies.  When coupled with the other networks in our framework, it quickly outpaces these alternatives.  We additionally demonstrate that the {\sc NRMN}s track objects well over time, due to the accuracy of the dense motion flow fields from the {\sc LFN}s.  Our framework does not lose contextual focus on a target unless it is predominantly out of view.  As well, we illustrate that zero-shot learning to new target types is possible with a modified, pre-trained {\sc NRMN}.  We achieve state-of-the-art performance compared to many existing zero-shot methodologies.  Our {\sc NRMN}s perform either similarly to or better than existing deep-network approaches for FLS imagery in the literature despite the former using orders of magnitude fewer training samples.

\subsection*{\small{\sf{\textbf{2.$\;\;\;$Comparisons}}}}\addtocounter{section}{1}

Using network architectures for target analysis in imaging sonar is not new \cite{StewartWK-jour1994a,MichalopoulouZH-jour1995a,ChakrabortyB-jour2003a,PerrySW-jour2004a,SongY-jour2021a}.  Many of these efforts have focused on the side-scan case, though, not the forward-looking case.

Some of the early work on the use of deep networks for side-scan-based target analysis was conducted by Denos et al. \cite{DenosK-conf2017a}, Gebhardt et al. \cite{GebhardtD-conf2017a}, and McKay et al. \cite{McKayJ-conf2017a}, among others \cite{WilliamsDP-conf2018a}.  Many of these authors relied on {\sc VGG}-inspired convolutional networks to iteratively extract and refine discriminative details about potential targets in a scene.  More recently, Yu et al. \cite{YuY-conf2018a} considered using deep, sparse autoencoder networks to classify passive-sonar targets.  Instead of directly processing the raw signatures, the authors performed a wavelet-packet-component-energy decomposition, which they claimed would provide some insensitivity to noise.  

Other uses of deep convolutional architectures have been investigated.  Song et al. \cite{SongY-conf2017a} proposed a hybrid convolutional-network and Markov-random-field model for side-scan sonar segmentation.  K\"{o}hntopp et al. \cite{KohntoppD-conf2017a} compared the performance of a reduced-complexity {\sc AlexNet} architecture against conventional attributes like texture descriptors \cite{LianantonakisM-jour2007a} and lacunarity \cite{WilliamsDP-jour2015b}.  They demonstrated that convolutional networks outperformed pre-specified-feature approaches, which they largely attributed to both the networks' spatial invariance properties and its ability to tailor the convolutional kernels to the samples.  In \cite{JinL-conf2017a}, Jin and Liang advocated using a pre-trained {\sc AlexNet} for recognizing various objects in underwater imagery.

Only a modest amount of work has been done for deep-network-based target analysis in forward-looking imaging sonar.  Much of it relies on applying pre-established convolutional architectures.  One of the earliest approaches was due to Valdenegro-Toro \cite{Valdenegro-ToroM-conf2016a,Valdenegro-ToroM-conf2016b}.  He considered a range of deep and shallow networks for recognizing clutter objects in forward-looking sonar.  He demonstrated that the former greatly outperformed the latter, regardless of the complexities of the networks.  Improvements were also witnessed over conventional template-matching schemes \cite{FandosR-jour2014a}.  Kim et al. \cite{KimJ-conf2016b,KimJ-conf2016a} leveraged a {\sc DarkNet} architecture for single-class mobile target detection, while Ye et al. \cite{YeX-conf2018a} utilized hybrid {\sc FCN}-{\sc Siamese} networks for the same task.  Kvasi\'{c} et al. \cite{KvasicI-conf2019a} analyzed the performance of {\sc DarkNet}-based {\sc YOLO} and {\sc TinyYOLO} networks for detecting and tracking mobile targets from a single class.  Fuchs et al. \cite{FuchsLR-conf2018a} used {\sc ResNet}s to discriminate between multiple target types.  They showed that pre-training on natural imagery vastly reduced the amount of sonar imagery required to yield good performance, despite that the first-layer receptive fields can differ greatly from those learned for the former.  Jin et al. \cite{JinL-jour2019a} proposed {\sc EchoNets}, fully convolutional networks for multi-class transfer learning and target recognition.

Here, we consider a novel deep-network framework for target feature extraction, detection and classification, and temporal label consensus in forward-looking sonar imagery.  Compared to the above approaches, ours is unique due to the use of content-addressable memories for efficiently storing invariant, multi-class target representations.  Such memories, when coupled with robust distractor removal, facilitate training and leveraging small-scale networks that can generalize well to new target type from few labeled samples.  No longer is the overall target recognition capacity of a network driven solely by complex, hierarchical feature extraction and manipulation.  Rather, it is a byproduct of storing maximally-distinct, mostly-distractor-free class instances and conducting relationship-based memory searches.  Both properties enable our framework to achieve good recognition performance when using only a fraction of the suggested sample size proposed by Valdenegro-Toro \cite{Valdenegro-ToroM-conf2017a}.  Ten to thirty instances of a target class is often sufficient to perform similarly to conventional deep networks, provided that the seafloor and other distractors can be adequately removed.  When conducting zero-shot learning, no class-specific examples are needed.  The above referenced works for FLS imagery cannot readily replicate zero-shot-learning behaviors without significant modification.  

We empirically compare aspects of our framework against about fifty alternatives.  When discussing these experimental results, we provide detailed explanations for how our framework differs and why it behaves well.

\subsection*{\small{\sf{\textbf{3.$\;\;\;$Methodology}}}}\addtocounter{section}{1}

In what follows, we outline our tri-network framework for real-time, low-shot target detection and recognition.  We first cover the saliency network, {\sc DNSS}, used to identify target-like regions in FLS image streams (see section 3.1).  We then describe the memory-based matching network, {\sc NRMN}, that extracts and stores representations of targets (see section 3.2).  We explain how to effectively transform the query responses from the content-addressable memory to arrive at an open-set classification response.  Lastly, we discuss how to integrate pseudo-temporal constraints into the detection and recognition steps, without the overhead of recurrent network components.  We do this using flow-field-based {\sc LFN} networks (see section 3.3).  For each of these networks, we justify our design decisions.  We also provide brief comparisons against alternatives to highlight the effectiveness of our contributions.

\begin{figure*}
   \hspace{-0.25cm}\begin{tabular}{l}
      \hspace{0.475cm}\includegraphics[width=6.0in]{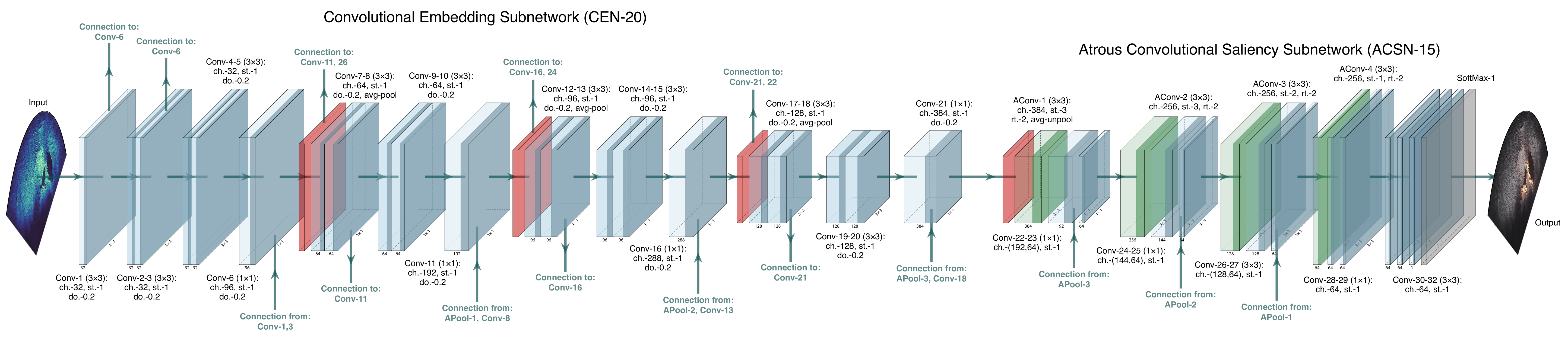}\vspace{-0.325cm}\\
      \hspace{3.275in}{\footnotesize (a)}\vspace{0.05cm}\\
      \hspace{-0.1cm}\includegraphics[width=6.55in]{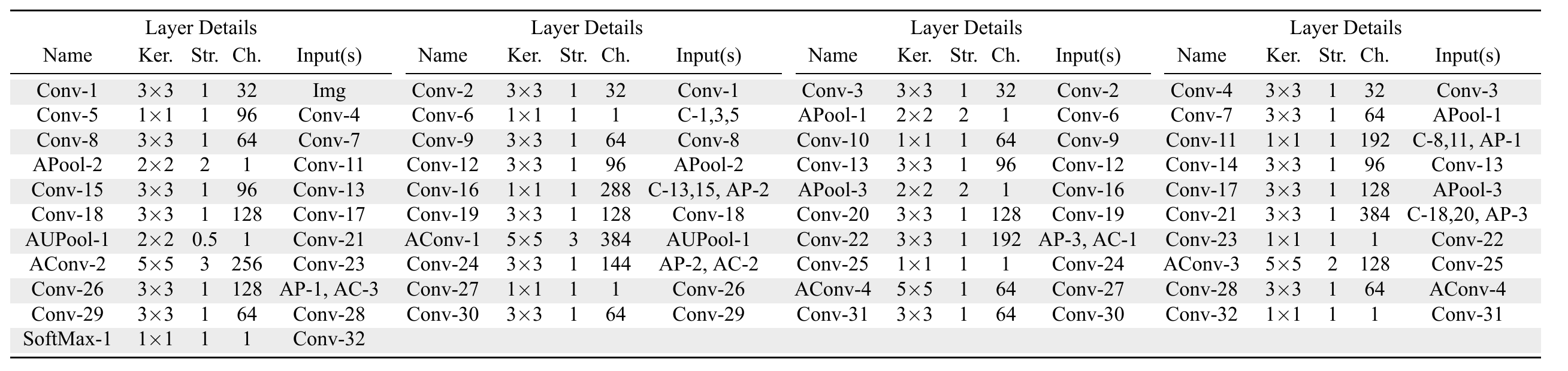}\vspace{-0.2cm}\\
      \hspace{3.275in}{\footnotesize (b)}
   \end{tabular}
   \vspace{-0.1cm}
   \caption[]{\fontdimen2\font=1.55pt\selectfont (a) A {\sc DenseNet}-inspired network, {\sc DNSS}, for fast multi-target focusing via saliency-based segmentation.  The {\sc DNSS} relies on two sub-networks, a {\sc CEN-20} and a {\sc ACSN-15}.  The former extracts progressively more rich features from a single FLS image that permit distinguishing between the seafloor, let alone distractor elements, and target-like objects.  The latter refines the feature representation and converts it into a probabilistic saliency map.  Skip connections are prevalent throughout both sub-networks to overcome information loss, in both directions, due to pooling.  Both sub-networks are pre-trained in a supervised fashion on natural imagery before being fit to FLS imagery.  For this diagram, convolutional layers are denoted using blue-colored blocks.  Average pooling layers are denoted using red blocks.  \`{A} trous spatial pyramid pooling layers are denoted using green blocks.  The softmax layer is denoted using a gray block.  (b) A tabular summary of the major network layers for the {\sc DNSS}.  In some instances, we shorten the names of various layers.\vspace{-0.4cm}}
   \label{fig:1}
\end{figure*}

\begin{figure*}[t!]
   \includegraphics[]{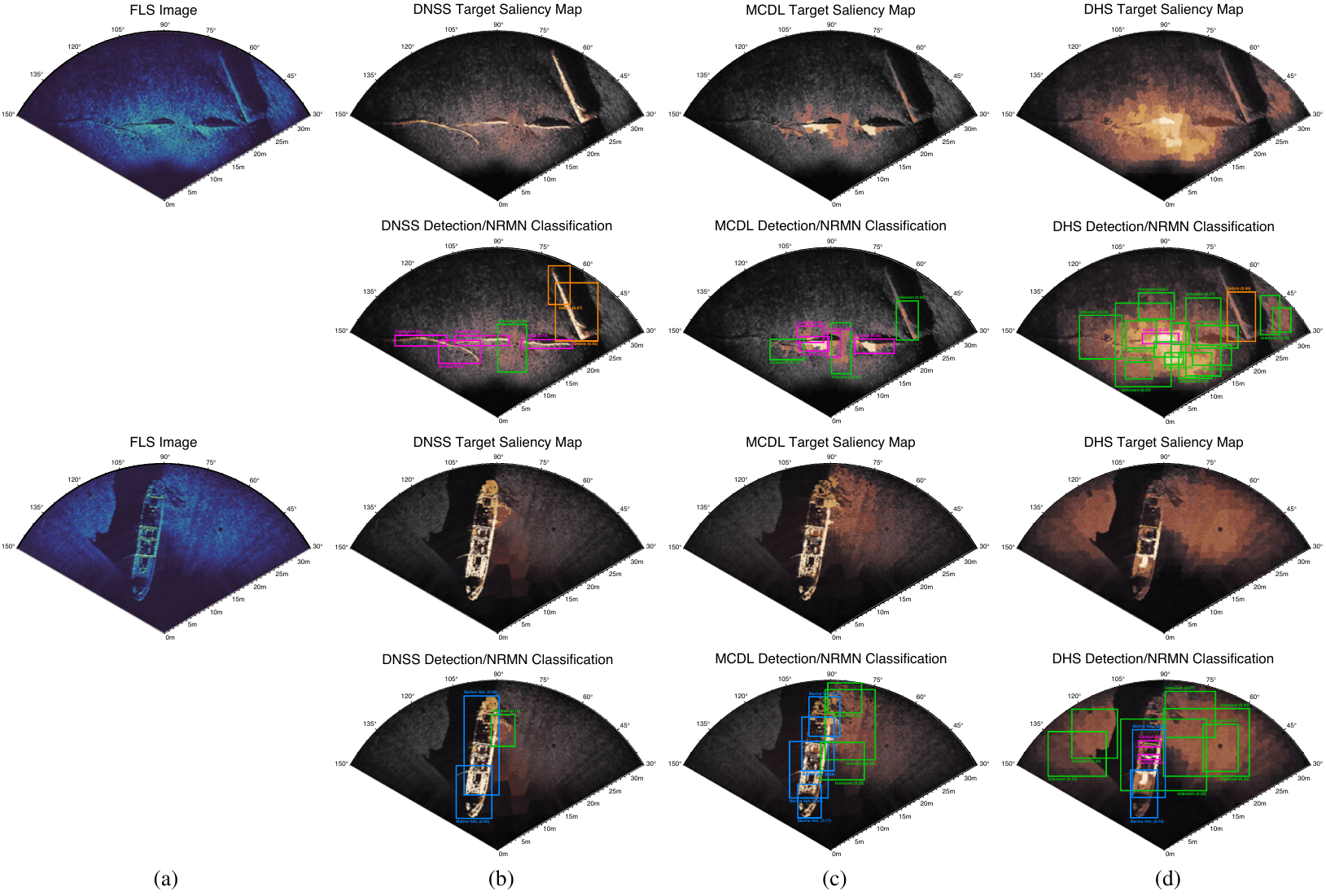}\vspace{0.1cm}
   \begin{tabular}{c}\hspace{0.1cm}\includegraphics[width=6.25in]{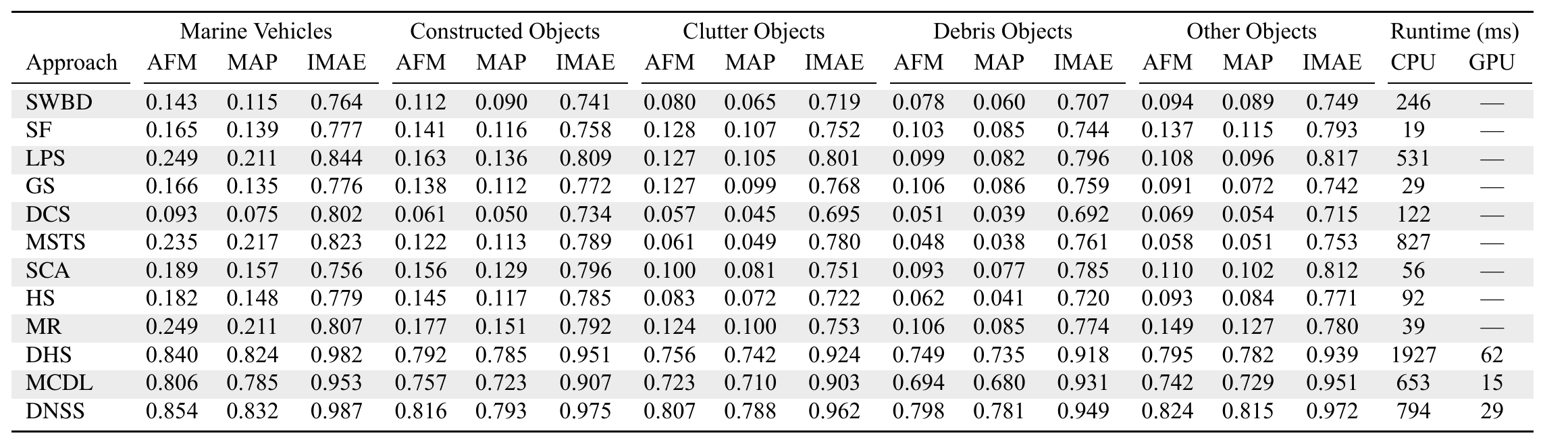}\vspace{-0.15cm}\\
   {\footnotesize (e)}$\;$
   \end{tabular}\vspace{-0.1cm}
   \caption[]{\fontdimen2\font=1.55pt\selectfont An overview of the importance of seafloor subtraction on low-shot learning performance with our {\sc NRMN}s.  Here, we consider a ten-shot, five-way learning strategy.  If the targets are not isolated well, then learning performance suffers when using only a small number of training samples.  There is too much variability in the imagery to adequately characterize just the targets.  (a) False-color FLS images of underwater scenes from the test set, which are displayed in a polar-wedge format.  These images have been adaptively despeckled and adaptively contrast equalized.  No temporal aggregation has been performed to enhance shape resolution.  (b)--(d) Target segmentation heatmaps for the corresponding underwater scene produced, respectively, by {\sc DNSS}, {\sc MCDL} \cite{LiuN-conf2016a}, and {\sc DHS} \cite{ZhouR-conf2015a}.  Each of these saliency-based segmentation approaches is deep-network-based.  Whiter colors denote higher segmentation confidence.  Beneath each heatmap, we show the corresponding detection boxes and target confidences, as returned by the {\sc NRMN}.  Each class is denoted using a distinct color.  (e) Table of saliency detection statistics for a 15000-image dataset when using the unsupervised, saliency-based segmentation schemes: SWBD \cite{ZhuW-conf2014a}, SF \cite{PerazziF-conf2012a}, LPS \cite{ZengY-conf2018a}, GS \cite{YangC-conf2012a}, DCS \cite{YangJ-conf2012a}, MSTS \cite{TuWC-conf2016a}, SCA \cite{QinY-conf2015a}, HS \cite{YanQ-conf2013a}, and MR \cite{YangC-conf2013a}.  Here, we use mean average precision (MAP), average f-measure (AFM) and inverse mean absolute error (IMAE) to quantify performance.  Higher values are better.  None of these unsupervised schemes can perform as well as deep networks, which limits their low-shot learning performance.  We also report runtime inference statistics, with lower values being better.  Such statistics indicate that {\sc DNSS} can be fielded in real-time-processing applications.\vspace{-0.4cm}}
   \label{fig:2}
\end{figure*}

\subsection*{\small{\sf{\textbf{3.1$\;\;\;${\sc DNSS}: Target Saliency Network}}}}

For zero- and low-shot learning to be conducive for FLS imagery, we have found it beneficial to remove superfluous, non-target-like details, which we refer to as distractors.  What remains is a limited set of candidate regions that can be quickly evaluated by small-scale networks to determine what types of targets, if any, exist.

Targets can have vastly different appearances, not all of which may be captured by the limited training samples.  It would not be appropriate to consider target-detection networks, like {\sc YOLO} and {\sc SSD}, for the purposes of isolating targets from distractors.  Such closed-set networks would fail for novel target types encountered in zero-shot-learning scenarios.  This would burden the matching network with adequately removing these distractors from the entire image, not just a small portion of it, which is not easy in low-shot-learning situations.  Instead, saliency-based object detection and segmentation should be used, as it will highlight visually conspicuous target-like regions, regardless of the target type.  Note that mostly bottom-up saliency \cite{GaoD-conf2007a,KleinDA-conf2011a,HouX-conf2007a}, not purely top-down saliency \cite{JuddT-conf2009a,JiaY-conf2013a,ShenX-conf2012a,LiuR-conf2014a}, is preferred.  We do not want a deep network to focus its attention on a limited set of target types.  Some top-down influences are needed, though, as the conventional center-surround assumption \cite{GaoD-conf2007a} does not necessarily apply for multi-target delineation in FLS imagery.

Our aim is thus to create a deep, convolutional network for iteratively constructing dense, pixel-level saliency masks that separate locally distinct visual regions.  We thus consider a small-scale, {\sc DenseNet}-inspired architecture (see \cref{fig:1}) for extracting contrast-based features, like color, intensity, orientation, and local gradients.  Such features are the foundation of texture.  Texture permits, for instance, differentiating between target-like regions and the seafloor, even in high-ambient-noise situations.  Targets are likely to be less visually homogeneous than the seafloor at local scales and hence more salient.

The {\sc DenseNet}-like network that we employ, {\sc DNSS}, is composed of four feature extraction blocks and five upsampling and refinement blocks.  We refer to these initial five blocks as the {\sc DNSS-CEN-20} and the remainder as {\sc DNSS-ACSN-15}.  As shown in \cref{fig:1}, the first block of the {\sc DNSS-CEN-20} contains five 32-channel convolutional layers that have 3$\times$3 neighborhoods and single strides (Conv-1--4).  Skip connection, with channel-wise feature concatenation, are used to propagate details from the first and third layers (Conv-1 and Conv-3) to the final part in this block (Conv-5), a 96 channel, 3$\times$3 convolutional layer.  Such connections overcome information loss and help prevent singularities caused by non-identifiability \cite{OrhanAE-conf2018a}.  The spatial dimension of the features are reduced by a 2$\times$2 average pooling kernel, with stride two.  The remaining blocks in the {\sc DNSS-ACSN-15} sub-network possess a similar structure, except that the number of channels increases to capture progressively more complex contrast features.  Leaky rectified-linear-unit activation functions are used for each convolutional layer to non-linearly transform the feature maps \cite{NairV-conf2010a}.  Batch normalization \cite{IoffeS-conf2015a} is applied to reduce internal covariate shift.

Our use of average pooling helps in extracting increasingly high-level semantic information that aids in saliency-based detection of target-like regions.  It encourages the network to identify the complete extent of the target \cite{ZhouB-conf2016a}, unlike other operations, such as max pooling.  This occurs because, when computing the map mean by average pooling, the response is maximized by finding all discriminative parts of an object.  Those regions with low activations are ignored and used to reduce the output dimensionality of the particular map.  Moreover, average pooling tends to remove speckle noise naturally, thereby improving saliency segmentation performance for FLS imagery than when max pooling is used.  Each pooling layer leads to a reduced feature map size, though, posing serious challenges when forming a full-resolution solution.  While recurrent-deconvolutional layers can progressively increase the segmentation-map resolution, they are highly parameter intensive and hence require great amounts of training data.  They also adversely impact sample throughput rates.

The \`{a}-trous convolution layer was proposed to resolve the contradictory requirements between large feature map resolution and large receptive fields in a parameter-efficient way \cite{ChenLC-jour2018a}.  In the {\sc DNSS-ACSN-15} sub-network, we use a form of it, \`{a}-trous spatial pyramid pooling, to upsample and refine a saliency-based segmentation map.  More specifically, in the {\sc DNSS-ACSN-15}, the first block contains a 2$\times$2 average unpooling layer with a half stride.  This is followed by a 5$\times$5 \`{a}-trous deconvolution layer (Aconv1), with a dilation rate of 3, and a 192-channel, 3$\times$3 convolutional layer (Conv-22), and a 1$\times$1 convolution layer (Conv-23), with single strides.  The first convolutional layer (Conv-22) integrates earlier-stage features to enhance the segmentation map quality, while the second layer (Conv-23) in this pair flattens the response to create an intermediate saliency map.  The remaining blocks share a similar structure but omit the unpooling layer.  They instead rely on increasingly large dilation rates to handle target regions of different sizes.  Rates of 6 (Aconv2), 12 (Aconv3), 18 (Aconv4), and 24 (Aconv5) are employed.  We insert dense connections between the last convolution layers at each block and the \`{a}-trous deconvolution layers of subsequent blocks to further improve the upsampling quality by avoiding kernel degradation \cite{YangM-conf2018a}.

In the final block of the {\sc DNSS-ACSN-15} sub-network, we efficiently flatten the channels via a 1$\times$1 convolutional layer (Conv-32) before passing the result through a soft-max layer (SoftMax-1) to characterize the probability that a given pixel belongs to a salient target.  This yields a saliency segmentation map that is the same resolution as the input imagery.  This map can be further refined by the final stages of a deep parsing network \cite{LiuZ-conf2015a} to better align with potential target boundaries.  However, we have found it more effective, and faster, to pass the raw saliency map to a small-scale, pre-trained convolutional network to produce detection bounding boxes.  Local contextual features, like acoustic shadows, are retained when using bounding boxes, which usually aids in target recognition.

For our simulations, we pre-train the {\sc DNSS} network on natural-imagery datasets, like HKU-IS, PASCAL VOC, and DUT-OMRON, before fitting it to the FLS imagery.  We use ADAM-based back-propagation gradient descent with mini batches \cite{KingmaDP-conf2015a} and a balanced cross-entropy loss \cite{MukhotiJ-coll2020a}, which is computed per mini-batch sample,
\begin{equation*}
-\sum_{i=1}^n\sum_{j=1}^c \Bigg(\!\delta_{y_i = j}\alpha_1(1 \!-\! w_{i,j}^\theta)^{\gamma_1} \textnormal{log}(w_{i,j}^\theta) + \delta_{y_i\neq j}(1 \!-\! \alpha_1)(w_{i,j}^\theta)^{\gamma_1}\textnormal{log}(1 \!-\! w_{i,j}^\theta)\!\Bigg)\! + \kappa_1\|\theta\|^2_2.
\end{equation*}
Here, $w^\theta_{i,j} \!\in\! \mathbb{R}_{0,+} \!\backslash (1,\infty)$ is the probabilistic saliency score for the $i$th pixel and $j$th class of an FLS image from the\\ \noindent mini batch, while $y_i \!\in\! \mathbb{R}_{0,+} \!\backslash (1,\infty)$ is the corresponding ground-truth label; the saliency map is naturally parameterized by the network parameters $\theta$, where $\kappa_1 \!\in\! \mathbb{R}_{0,+}$ is the weight decay factor.  The hyperparameter $\gamma_1 \!\in\! \mathbb{R}_{0,+} $\\ \noindent controls the regularization amount.  Setting it too high can zero out the gradients.  However, higher values are preferred over those closer to zero; reference \cite{MukhotiJ-coll2020a} lists heuristics for choosing good values.  The other hyperparmeter, $\alpha_1 \!\in\! \mathbb{R}_{0,+} \!\backslash 0$, addresses class imbalancing.  It can be set using inverse class frequency.  Such a supervised-learning-\\ \noindent based loss typically exhibits better learning rates over alternatives \cite{ZhangZ-coll2018a}, since, among other reasons, the magnitude of the network parameter change is proportional to the error.  It also permits adaptive, sample-importance reshaping, thereby improving performance for rarely encountered targets.  

We provide preliminary simulation results in \cref{fig:2}.  These statistics indicate that this small-scale network significantly outperforms a variety of unsupervised, non-deep saliency schemes.  It is competitive against supervised, deep saliency networks that are larger and, potentially, extract more complex spatial features.  The {\sc DNSS} has a much higher throughput rate, though.  We refer readers to \cite{SledgeIJ-jour2020a} for detailed explanations of the poor performance obtained for these alternatives.  In many cases, it is due to their adoption of a purely bottom-up assumption that is not suited for the visual characteristics of sonar imagery.  Note that our network proposed in \cite{SledgeIJ-jour2020a}, the {\sc MB-CEDN}, could be applied with slight modifications to FLS imagery.  Preliminary experimental results indicate that it yields qualitatively improved saliency maps compared to the {\sc DNSS}.  However, the {\sc MB-CEDN} is too computationally intensive to be fielded in real-time, high-throughput scenarios.  It uses many times more convolution filters.  {\sc DNSS}s are better suited for these settings due to their comparatively compact size.

\subsection*{\small{\sf{\textbf{3.2$\;\;\;${\sc NRMN}: Memory Matching Network}}}}

Once distractors have been removed, the target-like regions within the bounding boxes can be classified.  Although we could use a standard convolutional network for this purpose, we would prefer to have one that can generalize well from only a few or even no class-specific samples.  Manual data annotation requirements would also be significantly reduced for this second type of network.

\begin{figure*}
   \hspace{-0.25cm}\begin{tabular}{l}
      \hspace{0.075cm}\includegraphics[width=6.35in]{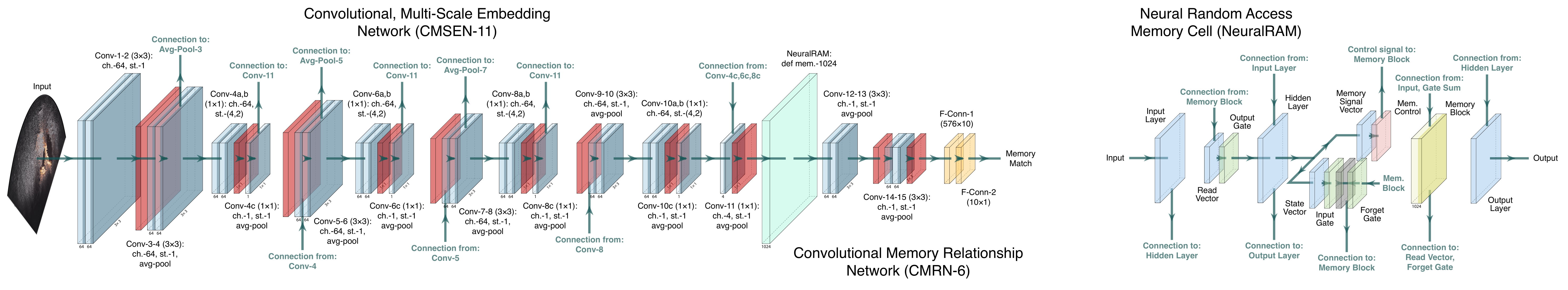}\vspace{-0.325cm}\\
      \hspace{2.15in}{\footnotesize (a)}\hspace{3.1in}{\footnotesize (b)}\vspace{0.05cm}\\
      \hspace{-0.1cm}\includegraphics[width=6.55in]{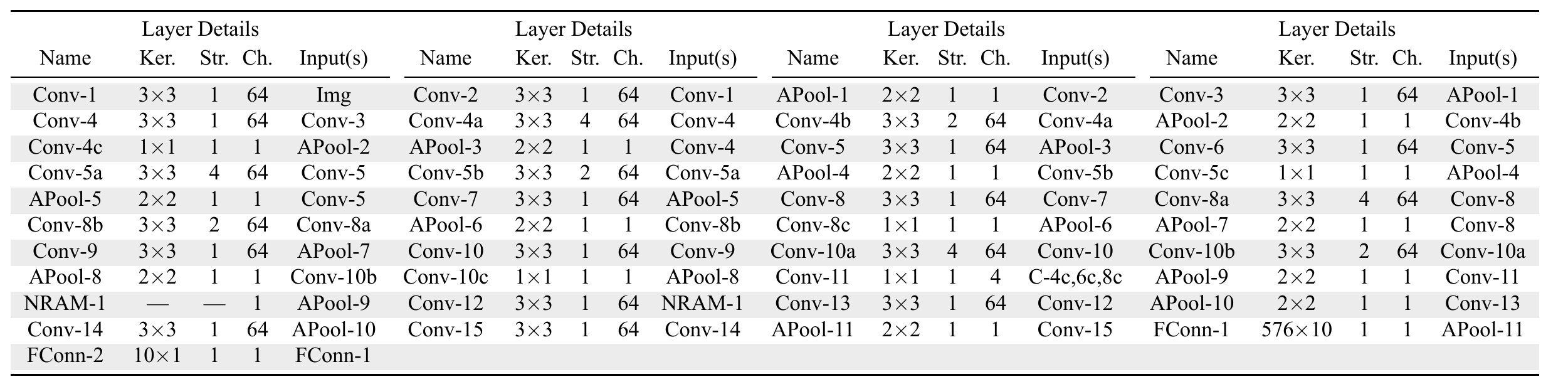}\vspace{-0.2cm}\\
      \hspace{3.275in}{\footnotesize (c)}
   \end{tabular}
   \vspace{-0.1cm}
   \caption[]{\fontdimen2\font=1.55pt\selectfont (a) A network diagram of a {\sc NRMN} for a given target candidate in a {\sc DNSS}-supplied saliency segmentation map.  The {\sc NRMN} relies on a series of convolutional blocks with residual skip connections to extract multi-scale classification features.  We refer to this part of the network as {\sc CMSEN-11}.  The {\sc CMSEN-11} sub-network is pre-trained in a supervised fashion on natural imagery before being fit to FLS imagery.  The output of this sub-network is a query feature vector that is used to access the memory banks of a content-addressable {\sc NeuralRAM} cell.  All relevant entries from the {\sc NRMN} are returned and fed into a convolutional relationship sub-network, which is referred to as {\sc CMRN-6}.  This sub-network specifies a deep similarity measure that discerns the most appropriate class for the target candidate in the saliency segmentation map.  For this diagram, convolutional layers are denoted using blue-colored blocks.  Average pooling layers are denoted using red blocks.  Fully connected layers are denoted using orange blocks.  The {\sc NeuralRAM} is denoted using a mint-colored block.  (b) An internal view of a recurrent {\sc NeuralRAM} cell.  Here, matrix/vector layers and gate responses are denoted using light blue and light green blocks, respectively.  The gate-response summation is denoted using a gray block.  The random-access memory and its associated read/write controller are, respectively, denoted using light yellow and light pink blocks.  (c) A tabular summary of the major network layers for {\sc NRMN}.  In some instances, we shorten the names of various layers.\vspace{-0.4cm}}
   \label{fig:3}
\end{figure*}

Here, we consider a memory-based matching network, {\sc NRMN}, for zero- and low-shot recognition (see \cref{fig:3}).  Memory-based networks are naturally suited for low-shot learning.  They can efficiently store and recall class-specific details.  The expressive power and discriminative capability of memory networks is tied to the number of memory entries, not necessarily the depth of the convolutional feature hierarchy.  This contrasts with standard convolutional networks whose expressive power is often a function of the number of layers.  These layers similarly encode class knowledge, albeit somewhat poorly, since this can only be done via the convolutional kernels.  Conventional networks must also simultaneously handle multiple classes through these kernels, which can be challenging, especially in sample-scarce settings.

The {\sc NRMN} has two sub-networks, {\sc NRMN-CMSEN-11} and {\sc NRMN-CMRN-6}.  The first is loosely based on {\sc ProtoNet}s \cite{SnellJ-coll2017a} and embeds the target-like regions from {\sc DNSS} into a low-dimensional feature space.  These features are used to interface with a content-addressable {\sc NeuralRAM}-based memory that stores representations for previous samples along with their class labels.  The second sub-network compares batches of test-set samples with the content of the neural memory to determine the most appropriate target class in an open-set fashion.

Like {\sc ProtoNet}s, the {\sc NRMN-CMSEN-11} sub-network is composed of four blocks of convolutional and pooling layers.  We keep the overall network size low by using single-stride convolutional layers with 64-channel, 3$\times$3 kernels (Conv-1--10). We, however, augment the standard {\sc ProtoNet} architecture so that it extracts multi-scale features, which characterizes differently sized targets well.  As shown in \cref{fig:3}, we connect three additional convolution layers to each block in {\sc NRMN-CMSEN-11} (Conv-1--4, Conv-5--6, Conv-7--8, Conv-9--10).  The first and second added layers use 3$\times$3 convolution kernels with 64 channels (Conv-4a/b, Conv-6/b, Conv-8a/b, Conv-10a/b).  The third added layer has a single-channel 1$\times$1 kernel, which is used to flatten the features (Conv-4c, Conv-6c, Conv-8c, Conv-10c).  To make the output feature maps of the four sets of extra convolutional layers have the same size, the strides are set to 4, 2, and 1, respectively.  Although the four produced feature maps are of the same size, they are computed using receptive fields with different sizes and hence represent contextual features at different scales.  We combine these four feature maps at the last convolutional layer of {\sc NRMN-CMSEN-11} (Conv-11) before one final downsampling step.

\begin{figure*}
   \includegraphics[]{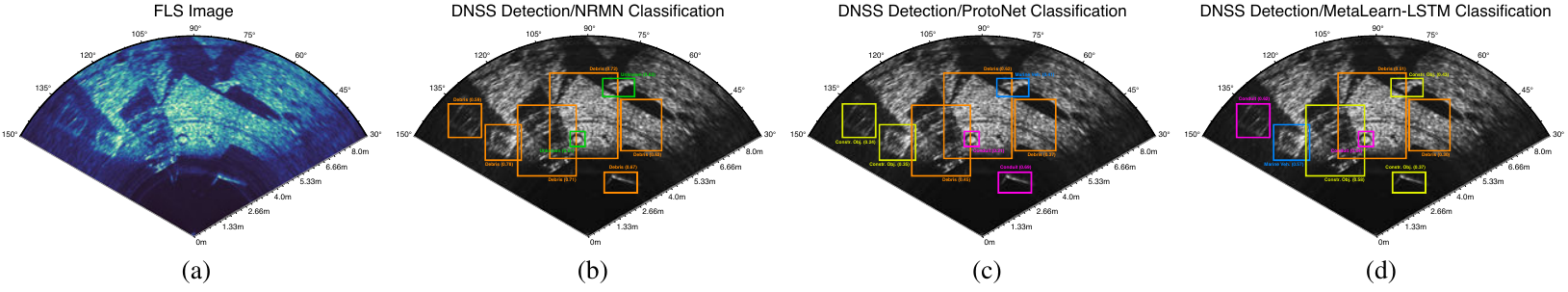}\vspace{-0.075cm}
   \begin{center}\begin{tabular}{c}\hspace{0.525cm}\includegraphics[width=4.85in]{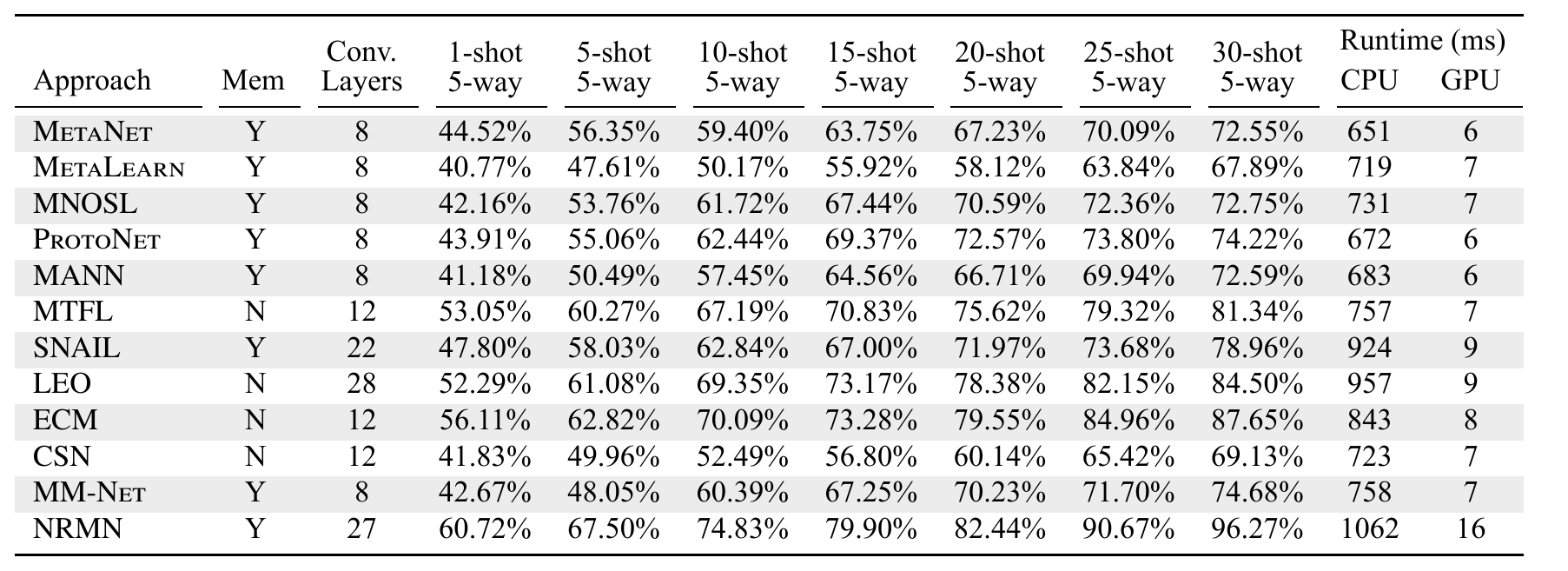}\vspace{-0.25cm}\\
   \end{tabular}
   \end{center}
   \hspace{8.5cm} {\footnotesize (e)}\vspace{-0.075cm}
   \caption[]{\fontdimen2\font=1.55pt\selectfont A comparison of memory and non-memory-based low-shot learners.  (a) A false-color FLS image of an underwater scene from the test set, which is displayed in a polar-wedge format.  This scene contains a variety of clutter objects, like a steel plate, a car tire, and an intermediate bulk container.  This image has been adaptively despeckled and adaptively contrast equalized.  No temporal aggregation has been performed to enhance shape resolution.  (b)--(d) Target detection boxes from {\sc DNSS} and classification responses produced, respectively, by {\sc NRMN}, {\sc ProtoNet} \cite{SnellJ-coll2017a}, and {\sc MetaLearn} \cite{RaviS-conf2017a} when considering 20-shot, 5-way learning.  Each class is denoted using a distinct color.  (e) Table of training and testing recognition percentages for a 5000-image dataset of large-sized targets when using {\sc MetaNet} \cite{MunkhdalaiT-conf2017a}, {\sc MNOSL} \cite{VinyalsO-coll2016a}, {\sc MANN} \cite{SantoroA-conf2016a}, {\sc MTFL} \cite{SunQ-conf2019a}, {\sc SNAIL} \cite{MishkinD-conf2018a}, {\sc LEO} \cite{RusuAA-conf2019a}, {\sc ECM} \cite{RavichandranA-conf2019a}, {\sc CSN} \cite{KochG-conf2015a}, and {\sc MM-Net} \cite{KaiserL-conf2017a}.  Higher values are better.  In the table, we identify which networks use memories.  We also specify the number of feature-extraction and modification convolution layers.  These alternate schemes often make mistakes because they not open-set classifiers.  We also provide runtime inference statistics, with lower values being better.  Such statistics indicate that memory-based networks can be fielded in real-time-processing applications.\vspace{-0.4cm}}
   \label{fig:4}
\end{figure*}

We force the final two average pooling layers (APool-5 and APool-7) to have a single stride, which prevents additional downsampling, and leads to a more modest eight times reduction of the input-imagery resolution.  Additionally, to maintain the same receptive-field size in the remaining layers (Conv-7--10), we apply an \`{a}-trous dilation operation to the corresponding filter kernels.  This operation allows for efficiently controlling the resolution of convolutional feature maps without the need to learn extra parameters, thereby improving training times.

The output of the {\sc NRMN-CMSEN-11} sub-network is a spatial feature map that is used to interface with a {\sc NeuralRAM} content-addressable memory (NRAM-1) \cite{WestonJ-conf2015a,GuicehreC-jour2018a}.  {\sc NeuralRAM} cells are recurrent, fully-differentiable networks that significantly generalize {\sc LSTM}s \cite{HochreiterS-coll1996a,HochreiterS-jour1997a}.  The former can store and recall arbitrary numbers of things simultaneously, whereas the latter can only memorize a single thing at a given instant \cite{MaY-jour2020a}.  Banks of {\sc LSTM}s would hence be needed to replicate the behavior of a {\sc NeuralRAM} cell, thereby complicating training.

A {\sc NeuralRAM} cell consists of a controller, which we take to be an {\sc LSTM} with a least-recently-used-access module \cite{SantoroA-conf2016a}.  We further enhance this controller by combining related class samples \cite{CaiQ-conf2018a}.  This controller interacts with the memory module of the {\sc NeuralRAM} using a number of reading and writing probes.  Memory encoding and retrieval in a {\sc NeuralRAM} cell is rapid, with either vector or matrix representations being either placed into or taken out of memory, potentially at every time-step.  This capability makes the {\sc NeuralRAM} uniquely suited for low-sample learning, as it is capable of long-term storage, through slow updates of its weights, and short-term storage, through its memory module.  If, during training, a {\sc NeuralRAM} can learn a general strategy for the types of representations it should place into memory and how it should later use these representations for predictions, then its speed can be exploited to make accurate predictions of samples, during testing, that the {\sc NRMN} has only seen once.

Previous implementations of {\sc NeuralRAM}-based matching networks for low-shot generalization relied on content-based querying of the memory module using fixed similarity measures.  The query sample would then be assigned to the same class as the memory entry with the highest similarity.  Such class-assignment process pre-supposes mostly linear separability of the stored representations, though, which is not likely to occur in practice.  We therefore rely on query-sample-sensitive transformations of the {\sc NeuralRAM} entries, which specifies a deep, non-parametric similarity measure.  This data-driven measure is implemented by the {\sc NRMN-CMRN-6}, which transforms the features in a way that more readily permits robust recognition, akin to \cite{HanX-conf2015a,ZagoruykoS-conf2015a}.  For each memory entry, we concatenate the stored representation (NRAM-1) and the current query-sample representation (APool-9) before processing by two sets of convolutional-pooling layers (Conv-12--13/APool-10 and Conv-14--15/APool-11).  The final feature map is flattened by two fully-connected layers (FConn-1--2) to determine the most appropriate memory entry and hence class.  As with the {\sc DNSS}, batch normalization is employed.  Rectified-linear-unit activations are used throughout except for the final fully-connected layer (FConn-2).  This layer uses an OpenMax activation \cite{BendaleA-conf2016a} to provide open-set recognition, which permits identifying that a potentially novel target type has been discovered.  Open-set recognition also offers a way for determine when more class-specific training examples are needed.

Zero-shot learning is analogous to one-shot learning in the sense that a single sample defines each additional class from a given set of base classes learned during training.  However, instead of supplying an image-based support set, these samples contain semantic information, such as textual descriptions of the visual attributes.  Due to the nature of this modality, we cannot process it in the same manner as image-based modalities in the {\sc NRMN}.  Rather, we use small-scale convolutional networks \cite{ChenJ-coll2018a} to embed this semantic content into a feature space.  The features can be combined with the result of the final layer from {\sc NRMN} (FConn-2), and subsequently transformed, to yield a recognition response for a given bounding box.

For our simulations, we pre-train the {\sc NRMN} on natural-imagery datasets, like MiniImageNet and ImageNet, before fitting it to the FLS imagery.  We clear the memory banks after pre-training.  We use ADAM-based back-propagation gradient descent with mini batches \cite{KingmaDP-conf2015a}.  A supervised-learning-based cross-entropy loss is employed for regressing the similarity score.  

We provide preliminary simulation results in \cref{fig:4} to illustrate that {\sc NRMN}s are competitive against other low-sample-learning networks, including those that are memory-based.  Here, we use detection boxes derived from the {\sc DNSS} saliency maps to provide region proposals to all of these networks.  We attribute our improvements to two key capabilities.  The first is the extraction of multi-scale features, which facilitates recognizing targets of vastly different sizes in the FLS imagery.  The second is the use of a deep similarity measure for querying the memory in an open-set fashion.  Absent these two traits, the {\sc NRMN}s yield almost identical results to other memory-based approaches.  We additionally note that without the candidate detection boxes from {\sc DNSS}, the performance of these other methods is incredibly poor.  They cannot generalize well in the presence of distractors, which we show in our comprehensive simulation study (see section 4).  They are unable to reconcile multiple target types in a single image and thus can only report the dominant class.  They also must make a closed-set classification decision, which is not appropriate in the presence of novel target types, let alone poorly modeled targets.

\subsection*{\small{\sf{\textbf{3.3$\;\;\;${\sc LFN}: Temporal Label Consistency Network}}}}

FLS image streams contain a great amount of temporal content that can be exploited for discriminability.  We, however, have opted to extract spatial, not spatio-temporal, features.  The former are far less parameter intensive and hence amenable to real-time processing.  We re-introduce some temporal information by inferring target motion fields throughout FLS image streams and using them to regularize the saliency segmentation process.  Doing this improves both detection and recognition accuracy, especially for small-scale targets, in the presence of strong noise.  This is because the uncertainty in the degraded image content is often significantly reduced by leveraging the segmentation solution for the content in regions from other, temporally-local FLS images that are better resolved.

\begin{figure*}
   \hspace{-0.25cm}\begin{tabular}{l}
      \hspace{0.0cm}\includegraphics[width=6.7in]{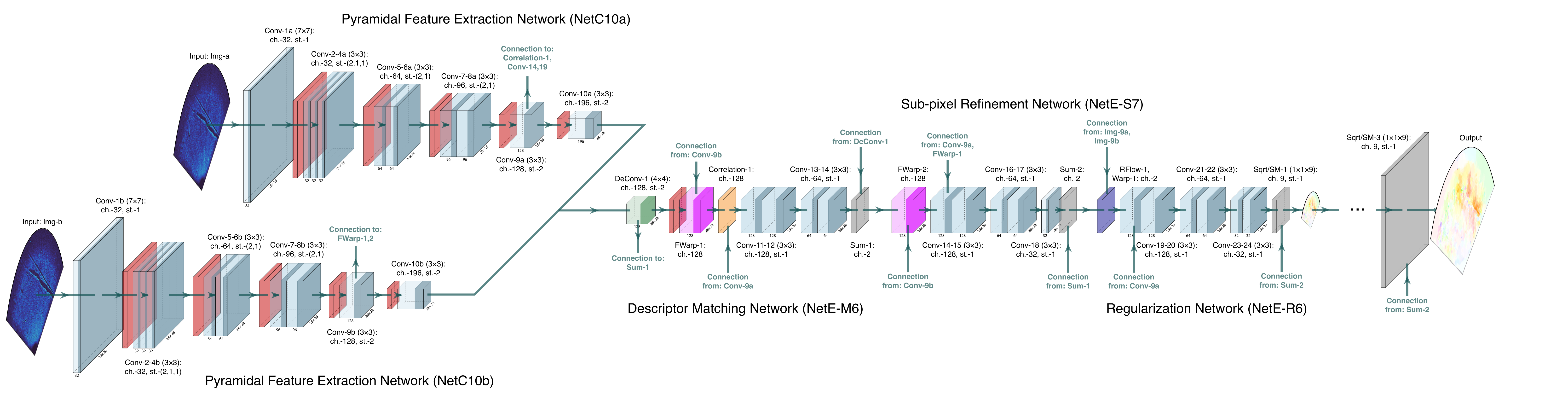}\vspace{-0.35cm}\\
      \hspace{3.275in}{\footnotesize (a)}\vspace{0.05cm}\\
      \hspace{-0.1cm}\includegraphics[width=6.55in]{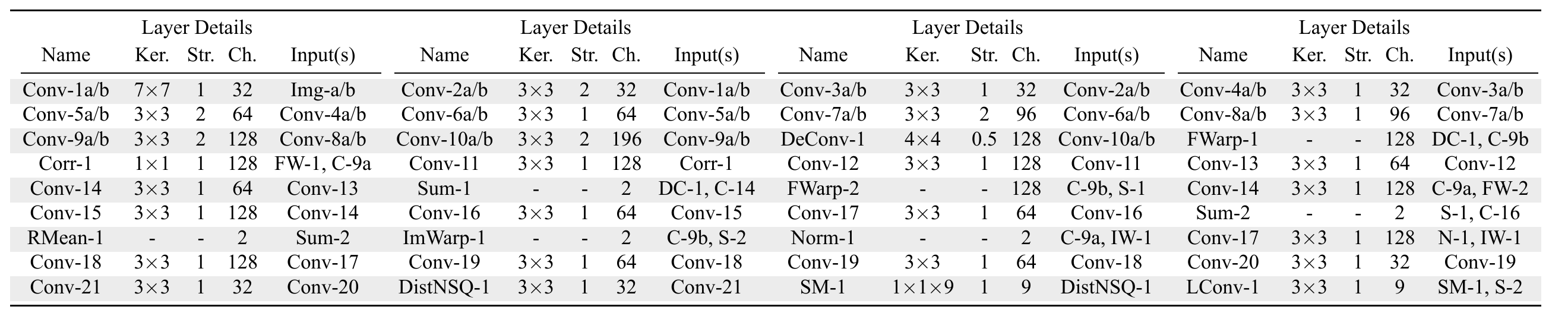}\vspace{-0.15cm}\\
      \hspace{3.275in}{\footnotesize (b)}
   \end{tabular}
   \vspace{-0.1cm}
   \caption[]{\fontdimen2\font=1.55pt\selectfont (a) A network diagram for the small-scale {\sc FlowNet}, {\sc LiteFlowNet}, used to estimate the dense, large-displacement scene correspondence between FLS image pairs.  The {\sc LiteFlowNet} relies on dual ten-layer encoders, referred to as {\sc NetC-10}s, to extract pyramidal, multi-scale features about images pairs that can be used to understand how scene content from one images changes to the next.  The filter weights for both encoders are tied together.  At each pyramidal level, a flow field is inferred from the high-level features.  This is performed by first matching descriptors according to a {\sc NetE-M6} sub-network.  The {\sc NetE-M6} sub-network iteratively constructs a volume of region alignment costs by aggregating short-range matching costs into a three-dimensional grid.  Since the cost volume is created by measuring pixel-by-pixel correlation, the resulting optical flow estimate from the previous pyramidal level is only accurate up to that level.  Sub-pixel refinement is necessary to extend the results to a new pyramidal level, which is performed by a {\sc NetE-S7} sub-network.  Lastly, to remove undesired artifacts and enhance flow vectors near target boundaries, a regularization network, referred to as {\sc NetE-R6}, is used.  This network relies on feature-driven local convolution to smooth the flow field in the interior of the target while preserving sharp discontinuities at the target edges.  Multiple {\sc NetE-M6}, {\sc NetE-S7}, and {\sc NetE-R6} sub-networks are cascaded to upsample and process the flow-field estimate; these additional networks are not shown in the above diagram.  For this diagram, convolutional and deconvolutional layers are denoted using, respectively, blue and green blocks.  Average pooling and unpooling layers are denoted using red blocks.  Softmax aggregation layers are denoted using gray blocks.  Correlation layers are denoted using orange blocks.  Flow warping and image warping layers are, respectively, represented by pink and dark blue blocks.  (b) A tabular summary of the major network layers.  In some instances, we shorten the names of various layers.\vspace{-0.4cm}}
   \label{fig:5}
\end{figure*}

\begin{figure*}
   \includegraphics[]{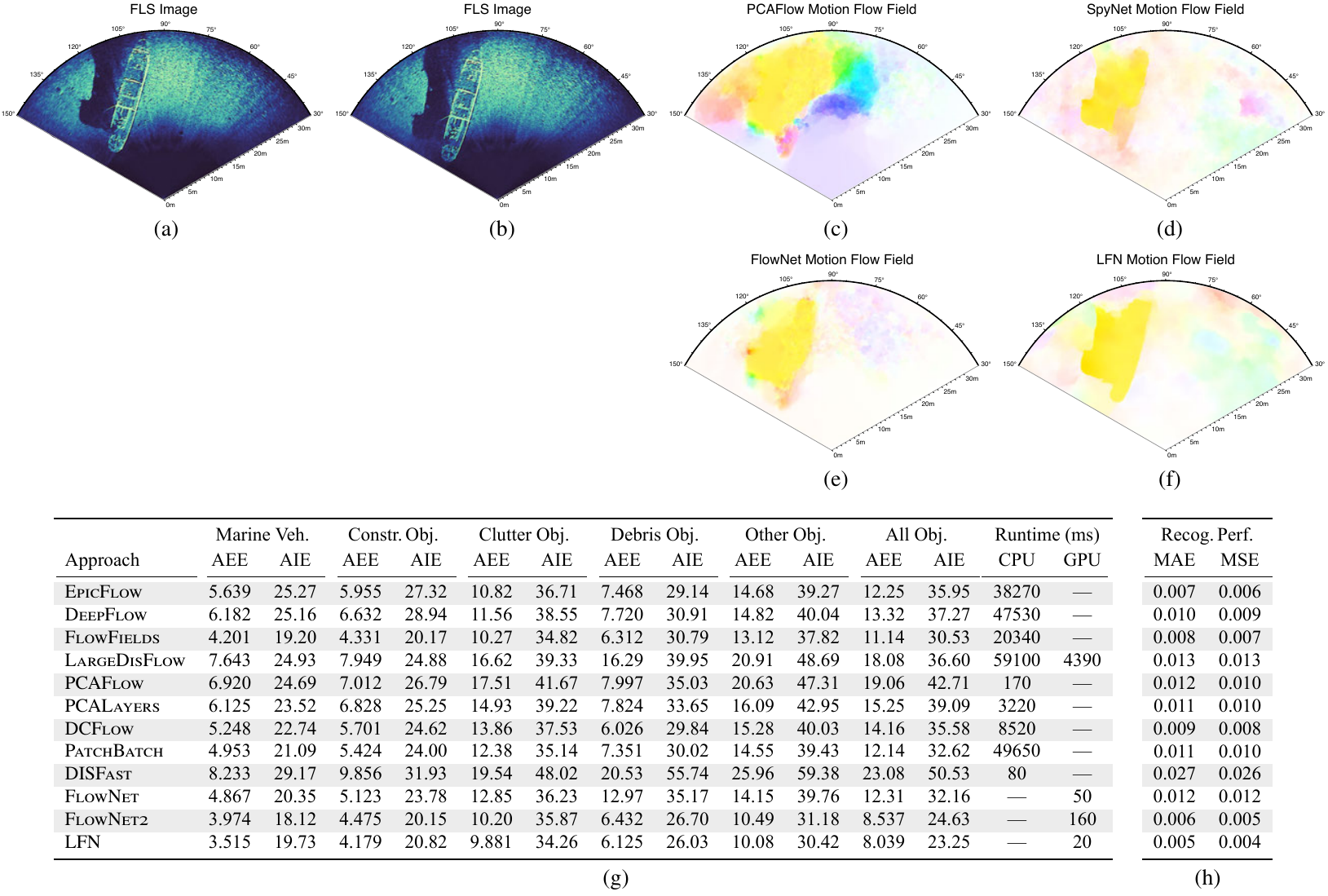}
   \caption[]{\fontdimen2\font=1.55pt\selectfont A comparison of motion flow fields for fast temporal label consensus.  (a)--(b) False-color FLS images of an underwater scene containing the \emph{Cape\! Fear} wreck, which are displayed in a polar-wedge format.  The sensing platform is heading from the stern of the shipwreck in the direction of the bow.  These images have been adaptively despeckled and adaptively contrast equalized.  No temporal aggregation has been performed to enhance shape resolution.  (c)--(f) Motion flow fields for this scene returned, respectively, by {\sc PCAFlow} \cite{WulffJ-conf2015a}, {\sc SpyNet} \cite{RanjanA-conf2017a}, {\sc FlowNet} \cite{DosovitskiyA-conf2015a}, and our {\sc LFN}.  The flow field for (f), qualitatively, is better for temporal label propagation than (c)--(e), as it focuses on the target and the cast acoustic shadow.  It extracts sharp target and shadow boundaries.  The methods used in (c)--(e) are sensitive to speckle noise that is prevalent in the FLS imagery.  They have difficulties in delineating target boundaries from motion.  Higher saturation denotes larger-magnitude displacements.  The color corresponds to the flow angle.  It is dictated by unwrapping the HSV colormap to the unit circle, where 0$^\circ$ corresponds to red, 90$^\circ$ to purple, 180$^\circ$ to cyan, and 270$^\circ$ to yellow.  (g) Table of flow-field statistics for a 50000-image dataset when using {\sc EpicFlow} \cite{RevaudJ-conf2015a}, {\sc DeepFlow} \cite{WeinzaepfelP-conf2013a}, {\sc FlowFields} \cite{BailerC-conf2015a}, {\sc LargeDisFlow} \cite{SundaramN-conf2010a}, {\sc PCAFlow} \cite{WulffJ-conf2015a}, {\sc PCALayers} \cite{WulffJ-conf2015a}, {\sc DCFlow} \cite{XuJ-conf2017a}, {\sc PatchBatch} \cite{GadotD-conf2016a}, {\sc DISFast} \cite{KroegerT-conf2016a}, {\sc FlowNet} \cite{DosovitskiyA-conf2015a}, and {\sc FlowNet2} \cite{IlgE-conf2017a}.  Here, we use the average endpoint error (AEE) and average interpolation error (AIE) to quantify performance.  Lower values are better.  (h) Table of label consensus statistics in terms of mean absolute error (MAE) and mean squared error (MSE).  Lower values are better.  While some of these methods do well, their inference can be slow, which prevents real-time detection and recognition.\vspace{-0.4cm}}
   \label{fig:6}
\end{figure*}

Here, we use modified, small-scale version \cite{HuiTW-conf2018a} of the {\sc FlowNet} architecture \cite{DosovitskiyA-conf2015a,IlgE-conf2017a}, {\sc LFN} (see \cref{fig:5}).  Such a network extracts features from the FLS imagery that iteratively inform a dense correspondence map between image pairs.  This is done by two sub-networks.  The first, {\sc LFN-NetC10}, converts any given image pair into pyramids of multi-scale features.  Several stacks of {\sc LFN-NetE19} sub-networks are employed to perform cascaded flow inference and regularization that result in coarse-to-fine flow fields.  Such motion flow fields provide a non-linear warping of sonar image content.  Previously computed segmentation maps can be transformed along those same fields, allowing that solution to be propagated to initialize the mask for a new FLS image.

As shown in \cref{fig:5}, {\sc LFN-NetC10} is a ten-layer, dual-stream network that transforms a pair of sonar images into pyramids of multi-scale features.  This is done through a series of multi-stride convolutional layers.  The first layer (Conv-1a/b) relies on a 7$\times$7 kernel with 32 channels and a single stride.  This is followed by three 3$\times$3 convolutional layers with 32 channels each (Conv-2--4a/b).  The of these three layers (Conv-2a/b) has stride two and the remainder (Conv-3--4a/b) have a single stride.  Four convolutional layers then follow (Conv-5--7a/b), each of which again rely on 3$\times$3 kernels.  They have 32, 64, 96, and 96 channels, respectively, along with strides of two, one, two, and one.  Lastly, two 3$\times$3 convolutional layers (Conv-9-10a/b), with dual strides, are used.  They have 96 and 128 channels, respectively.  Leaky rectified-linear units are inserted after every convolutional layer.  At each pyramid level of {\sc LFN-NetC10}, a flow field can be inferred from the high-level features of both images, versus the images themselves, using a differentiable bilinear interpolation.  This makes the {\sc LFN} robust to large-scale displacements.  As the features progress through these {\sc LFN-NetC10} layers, their spatial resolution is reduced so as to capture increasingly prominent and sizeable spatial changes.  The network weights are shared across both {\sc LFN-NetC10} streams, which lowers the number of tunable parameters without noticeably impacting performance.

The {\sc LFN-NetE19} sub-networks operate on the {\sc LFN-NetC10}-derived features to perform pixel-by-pixel matching at progressively higher scales.  These matches are refined, eventually yielding a sub-pixel-accurate flow map.  This is done according to three processing blocks, which carry out descriptor matching, refinement, and regularization.  For the six-layer descriptor matching block, {\sc NetE-M6}, an initial 4$\times$4 deconvolution layer (DeConv-1) is used to spatially upsample the previous flow-field estimate by a factor of two.  This is followed by feature-warping (FWarp-1) and correlation layers (Correlation-1), which provide a point-point correspondence cost between images.  Four successive 3$\times$3 convolutional layers (Conv-11-14), with 128, 64, 32, and 2 channels are used to construct a residual flow from the cost volume.  The upsampled flow-field estimate and residual flow are summed to account for any changes at the particular current scale that could not be predicted solely through the deconvolution.  This typically yields accurate flow maps up to that scale.  The accuracy is improved further by the seven-layer sub-pixel refinement, {\sc NetE-S7}, network block.  In particular, a secondary residual flow field is computed via minimizing the feature-space distance between one image and an interpolated version of the second image.  Erroneous artifacts are hence de-emphasized when being passed to the next pyramid level.  The {\sc NetE-S7} block is composed of a feature-warping layer (FWarp-2) and four 3$\times$3 convolutional layers (Conv-14-17) with 128, 64, 32, and 2 channels.  Each convolutional layer has a single stride.  Leaky rectified-linear units are inserted after every convolutional layer.  An element-wise sum layer combines the residual flow field with the result from {\sc NetE-M6}.

Even with sub-pixel refinement, distortions and vague flow boundaries may still be present in the fields.  These issues can disrupt the resulting warped saliency segmentation maps.  We remove such details using feature-driven local convolution.  It adaptively smooths the flow field.  It is implemented by the eight-layer regularization block, {\sc LFN-NetE-R6}.  This regularization block acts as an averaging filter if the flow variation over a given patch has few discontinuities.  It also does not over-smooth the flow field across boundaries.  To realize this behavior, we define a feature-driven distance metric that estimates local flow variation using the pyramidal features from {\sc LFN-NetC10}, the inferred flow field from the previous {\sc LFN-NetE-S7} network, and an occlusion probability map.  First, we remove the mean from the inferred flow field and warp it with respect to a downsized version of the second input image.  We then apply a Frobenius norm to the difference of the color intensity values of this result and the color intensity values of a downsized version of the first input image.  This result (RFlow-1) is subsequently transformed by a bank of five 3$\times$3 convolutional filters (Conv-19-24) with 128, 128, 64, 64, and 32 channels (Conv-17--21); they each have a single stride.  A normalized Boltzmann function is then applied.  Such an operation is defined using a 3$\times$3 convolutional-distance layer, with 9 channels, along with a negative square-root layer and a soft-max layer (Sqrt/SM-1).  A 3$\times$3 locally connected convolutional layer (LConv-1), with 2 channels, forms the final layer.  This layer adaptively constructs potentially unique soft-max-based filters for individual flow patches and convolves them with the inferred flow field from {\sc LFN-NetE-S7} to remove the undesired artifacts.  This yields a flow-field estimate at a given image resolution that is then progressively upsampled and processed by additional {\sc LFN-NetE} sub-networks.

Once the resolution of the flow field equals that of the input FLS images, then the saliency maps from previously processed images can be transferred along the vector field to form an initial saliency map for a new sonar image.  We find that considering the last five maps works well for both slow- and fast-moving sonar sensor platforms.  Using too many previous saliency maps can cause issues when the platform velocity is high and hence image content changes drastically over short time scales.  Structured local predictors \cite{BuloSR-conf2012a} are employed to facilitate label consensus amongst multiple images by taking into account local neighborhood appearance, relative position, and warped segmentation labels.  The transferred segmentation estimate is then concatenated with feature maps from the final block of the {\sc DNSS-ACSN-15} (Conv-30).  This provides regularization constraints for the saliency map construction of the currently considered FLS image, which generally improves the final map quality.  It is crucial for small-scale, fast-moving mobile targets, which may sometimes appear as just ambient noise in a single FLS image.

For our simulations, we pre-trained a {\sc LFN} on natural-imagery datasets, like KITTI 2012 and KITTI 2015, before fitting it to the FLS imagery.  The network was trained in a stage-wise fashion.  We use the same data augmentation strategy, including noise injection, and training schedule as {\sc FlowNet2} \cite{IlgE-conf2017a}.  We employ a two-term training cost consisting of the average endpoint error and the average image interpolation error,
\begin{equation*}
\sum_{k=1}^m \alpha_{2,k} \Bigg(\!\sum_{i=1}^n (w^\theta_{k,i} \!-\! y_{k,i})^2 + \frac{|\gamma_2 \!-\! 2|}{\gamma_2}\Bigg(\!\Bigg(\!\frac{(x^{t}_{k,i}/\epsilon_k \!-\! x^{t-1}\!\circ\! w^\theta_{k,i}/\epsilon_k)^2}{|\gamma_2 \!-\! 2|} \!+\! 1\!\Bigg)^{\!\!\gamma_2/2} \!-\! 1\!\Bigg)\!\Bigg)\! + \kappa_2\|\theta\|^2_2.
\end{equation*}
Here, $w^\theta_{k,i} \!\in\! \mathbb{R}_{0,+} \!\backslash (1,\infty)$ is the predicted flow field for the $i$th pixel at the $k$th {\sc LFN} upsampling stage for an FLS\\ \noindent image from the mini batch, while $y_{k,i} \!\in\! \mathbb{R}_{0,+} \!\backslash (1,\infty)$ is the corresponding ground-truth flow field.  This flow field is appropriately downsampled to match the prediction resolution for each upsampling stage.  The predicted response is naturally parameterized by the network parameters $\theta$, where $\kappa_2 \!\in\! \mathbb{R}_{0,+}$ is the weight decay factor.  For the second term, $x_{k,i}^t$ represents the current FLS image while $x^{t-1} \!\circ\! w^\theta_{k,i}$ is the previous FLS image in a sequence that is warped according to the predicted flow field.  The hyperparameter $\gamma_2 \!\in\! \mathbb{R}_{0,+}\!\backslash 0$ controls the influence of outliers in this term, while $\epsilon_k \!\in\! \mathbb{R}_{0,+}\!\backslash 0$ is a small constant that acts as a scale parameter \cite{BarronJT-conf2019a}.  The hyperparameters $\alpha_{2,k} \!\in\! \mathbb{R}_{0,+}\!\backslash 0$ dictate the importance of each upsampling stage.  We zero the contribution of the first term when fitting the {\sc LFN} to the FLS imagery, since we do not currently have dense ground-truth flow fields, just sparse ones.  Both errors are standard for flow estimation.

We provide preliminary simulation results in \cref{fig:6}.  They indicate that the {\sc LFN} significantly outperforms alternate deep networks while providing a much higher throughput.  The {\sc LFN} provides qualitatively better target-focused flow fields that facilitate improved, quantitative label transfer for low-shot learning in FLS imagery.  This improvement stems from the feature-driven local convolution.  This convolution acts as a regularization term to smooth the flow field, which mitigates effects of acoustic noise, while maintaining sharp flow boundaries.  It also arises from the use of descriptor matching, which addresses large-displacement motions, and sub-pixel refinement, which yields detail-preserving flows in the presence of noise.

\setcounter{figure}{0}
\subsection*{\small{\sf{\textbf{4.$\;\;\;$Simulations}}}}\addtocounter{section}{1}

We broadly assess the capability of our framework for detecting and recognizing targets in real-world FLS imagery (see section 4.1).  We demonstrate that {\sc DNSS}s can reliably isolate targets in an aspect-independent manner, regardless of their appearance.  This facilitates multi-target, zero- and low-shot generalization (see section 4.2).  Our framework hence compares well against {\sc YOLOv3} and {\sc TinyYOLOv3}, let alone other networks previously considered for FLS imagery.  This is despite our framework using far fewer domain-specific samples.  Lastly, we compare the {\sc DNSS}-guided {\sc NRMN}s against other zero- and low-shot learners (see section 4.2).  We show that the latter struggle for multi-target recognition, which reduces their effectiveness for this imaging-sonar modality.

\subsection*{\small{\sf{\textbf{4.1.$\;\;\;$Simulation Data}}}}

For our simulations, we utilize dual-band, high-resolution FLS sensors, including the Teledyne-RESON 7130, Kongsberg Flexview, and Oculus M750d and M1200d.  We utilize the high-frequency arrays, which are, respectively, 635~kHz, 1200~kHz, along with 1200~kHz and 2100~kHz.  At these frequencies, the sensing range is limited by absorption losses.  The maximum ranges have been observed to be approximately 10--200 m, depending on the sensor, in shallow-water environments.  The along-range resolutions of these sensors are, respectively, 25~mm, 10~mm, and 2.5~mm.  Consequently, image quality increases as the distance to observed targets decreases.  The horizontal beamwidths range from 60--70$^\circ$, for the M750d and M1200d, to 120--140$^\circ$, for the remaining sensors.  The vertical beamwidths are 10$^\circ$--33$^\circ$.  The horizontal angular resolutions are 0.6$^\circ$--0.95$^\circ$.  The update rates are anywhere from 10--40 Hz, implying that any real-time framework should process at least ten frames per second, which ours does.

These sensors were mounted on REMUS, MARV, and SeeByte unmanned undersea vehicles.  Each vehicle operated in a variety of littoral and oceanic environments throughout the continental United States.  Data collection spanned multiple years.  For each sensor, we annotated, respectively, about 15000, 60000, and 72000 FLS images that captured a variety of differently-scaled targets.  Relatively small targets, compared to the FLS image size, dominate the Teledyne-RESON dataset.  It is a challenging dataset.  The remaining two datasets contain targets that are comparatively larger.  These are somewhat easier datasets.  We partitioned the observed targets into broad classes.  We also provided hierarchical labels for the present sub-classes.  The image streams for various targets were segmented in time to localize when targets came into view.  The original datasets were composed of many times more images, but they were largely devoid of both targets and unique seafloor features.  Using them did little to enhance training.

\subsection*{\small{\sf{\textbf{4.2.$\;\;\;$Simulation Results and Discussions}}}}

\subsection*{\small{\sf{\textbf{4.2.1.$\;\;\;$Comparative Deep-Network Performance}}}}

We first compare the performance of {\sc DNSS}-guided {\sc NRMN}s with conventional deep networks, including many that have been previously used for processing FLS image streams.

\vspace{0.15cm}\noindent {\small{\sf{\textbf{Simulation Protocols.}}}} For each of the comparative deep networks, we rely on their standard backbone architectures for feature extraction.  If multiple backbones are compared in a given reference, then we select the one with better performance.  We pre-train each of these deep detectors on PASCAL VOC 2007 and 2012, along with COCO.  We then fit the networks to the FLS datasets.  Unless otherwise stated in the references, we employ ADAM-based back-propagation gradient descent with mini batches for training.  We use hyperparameter values listed in the references; there are too many to reproduce here, but they will be provided as an online supplement upon acceptance of our paper.

For each simulation of these alternate networks, we randomly split the data into training, testing, and validation sets with ratios of 80$\%$, 10$\%$, and 10$\%$, respectively.  We set aside 5\% of the samples from all datasets to be used solely as a test set.  We ensure that each class is well represented in each of these three sets.  Pre-training on the natural-image datasets is terminated once the network loss on the validation set monotonically increased for five consecutive epochs, where an epoch is a full presentation of the training set.  This preempts overfitting.  Training could also be terminated if no improvement on the training set was observed across two epochs and the validation-set change was below some threshold.  Learning then commenced on the FLS imagery and was terminated in the same fashion.  We choose the models with the lowest validation-set loss for a given run when computing statistics.

For our framework, we assess the visual similarity of the FLS imagery \cite{SongHO-conf2016a}.  We randomly sample 50\% of the training set from the most unique images of a given class.  The remaining 50\% is composed of those images that are either relatively or highly similar to each other.  This improves generalization for challenging scenes.  Out of the remaining samples, 10\% is taken for a validation set and the remaining 90\% is used as a test set.  We set aside 5\% of the samples from all datasets to be used solely as a test set.  We again select the models with the lowest validation-set loss for a given run when computing statistics.

We report our statistics on the combined test and validation sets.  All statistics are averaged over across five Monte Carlo runs.  We exclude runs where learning stalled and hence terminated early, which could be due to either poor parameter initializations or other factors.

\begin{figure*}
   \includegraphics[]{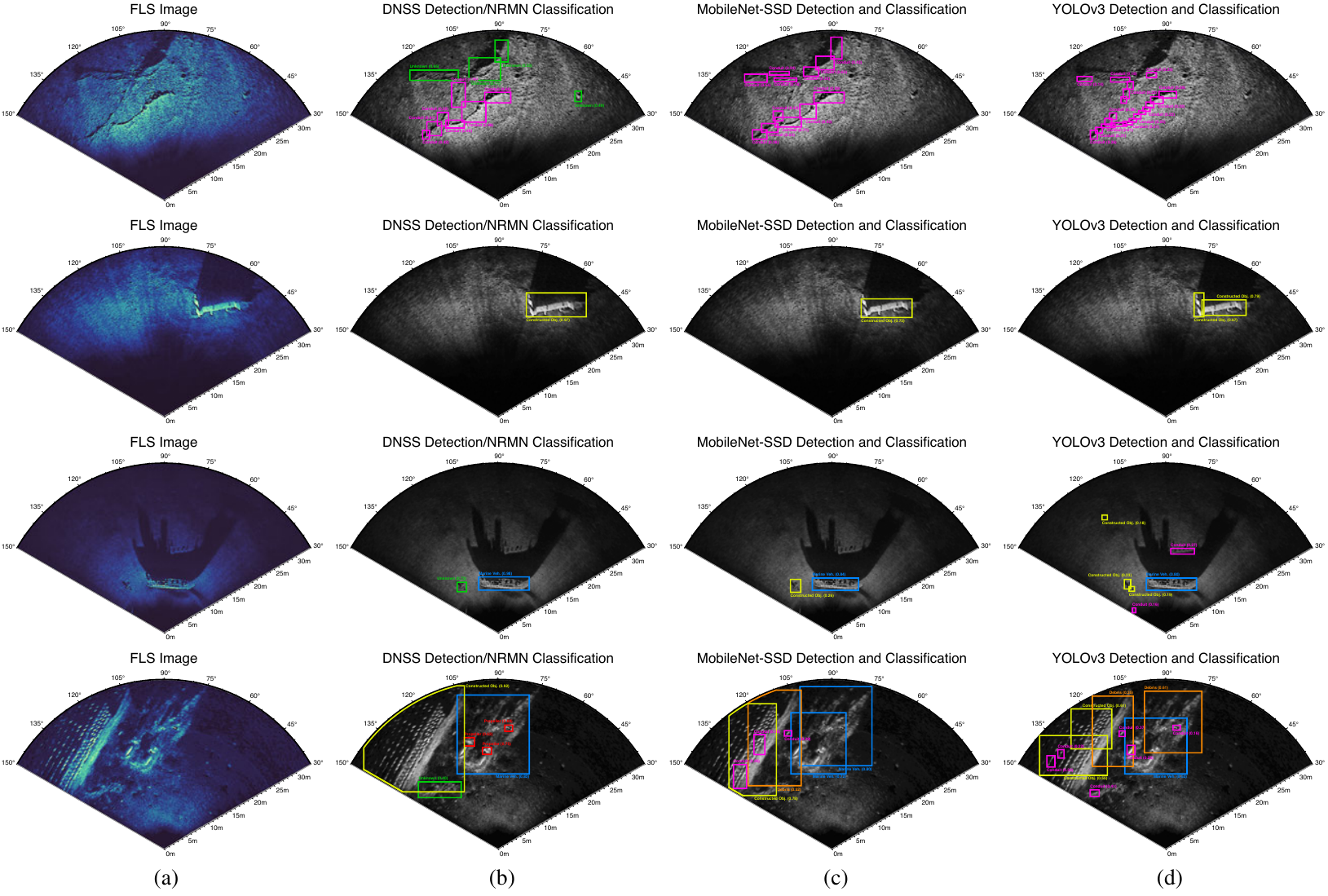}\vspace{0.15cm}
   \begin{tabular}{l}\hspace{-0.3cm}\includegraphics[width=4.6in]{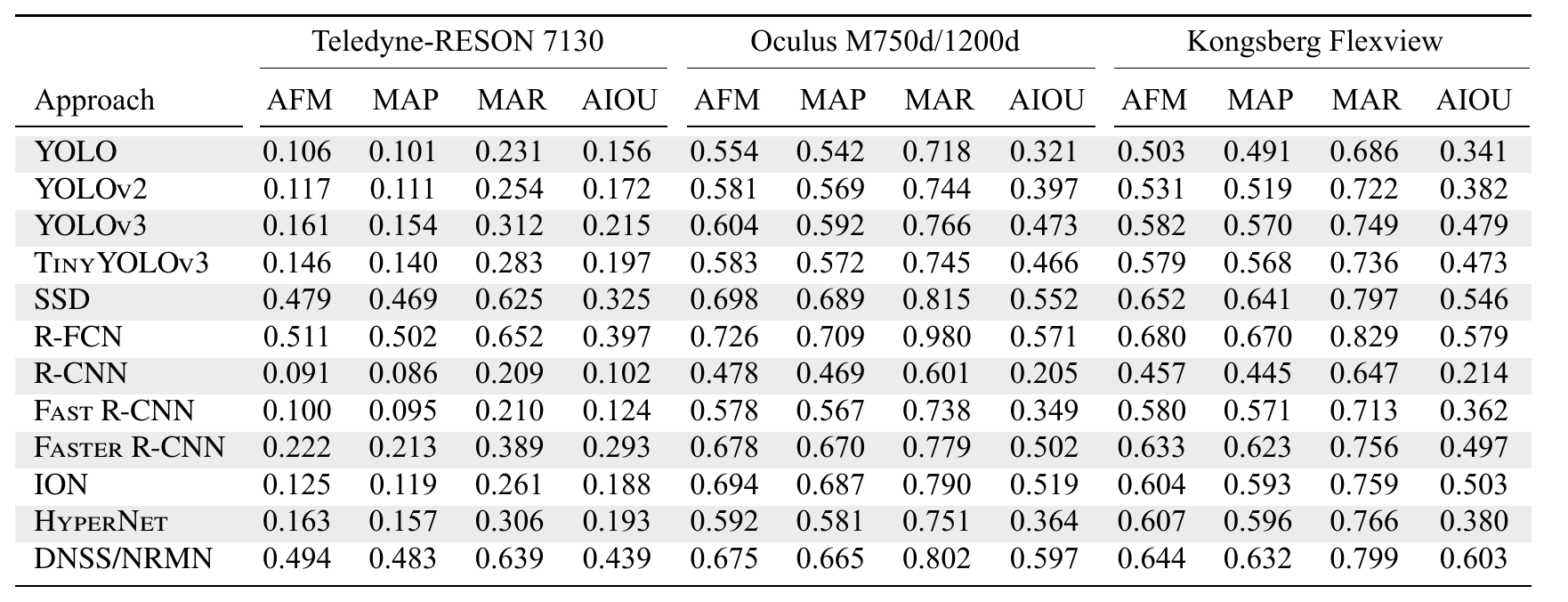}\includegraphics[width=2.15in]{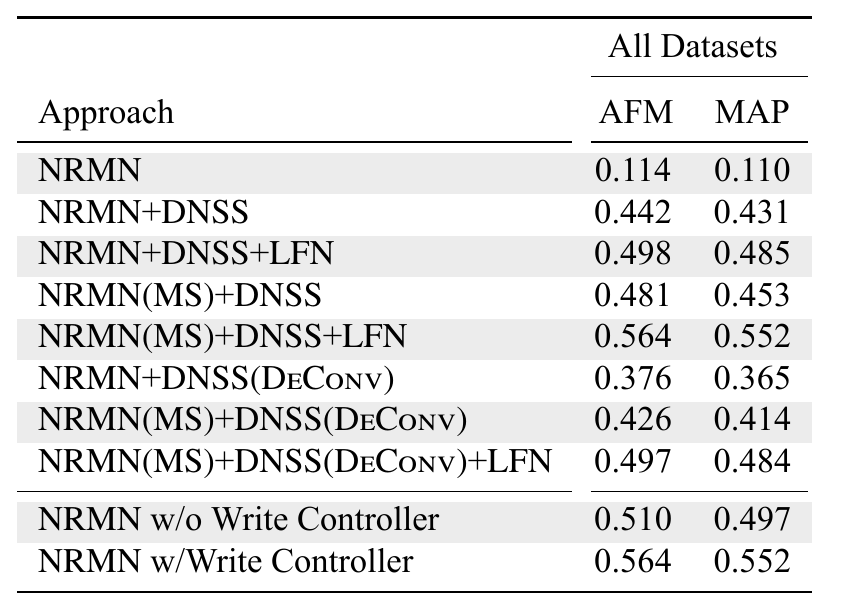}\vspace{-0.175cm}\\
   \hspace{5.5cm}{\footnotesize (e)} \hspace{7.95cm}{\footnotesize (f)}
   \end{tabular}\vspace{-0.0cm}
   \caption[]{\fontdimen2\font=1.55pt\selectfont A comparison of {\sc DNSS}-guided {\sc NRMN}s against other combined detection-recognition networks.  Here, we consider a ten-shot, multi-way learning strategy for the {\sc NRMN}s.  (a) False-color FLS images of underwater scenes from the test set, which are in a polar-wedge format.  These images have been adaptively despeckled and adaptively contrast equalized.  No temporal aggregation has been performed to enhance shape resolution.  (b)--(d) Target detection boxes and target confidences, as returned, respectively, by the {\sc NRMN}, {\sc MobileNet-SSD} and {\sc YOLOv3} deep networks.  Each class is denoted using a distinct color.  (e) Table of combined labeled detection statistics for various comparative deep networks, which includes {\sc YOLO} \cite{RedmonJ-conf2016a}, {\sc YOLOv2} \cite{RedmonJ-conf2017a}, {\sc YOLOv3}, {\sc TinyYOLOv3}, {\sc SSD} \cite{LiuW-conf2016b}, {\sc R-FCN} \cite{DaiJ-coll2016a}, {\sc R-CNN} \cite{vandeSandeKEA-conf2011a}, {\sc Fast R-CNN}  \cite{GirshickR-conf2015a}, {\sc Faster R-CNN} \cite{RenS-coll2015a}, {\sc ION} \cite{BellS-conf2016a}, and {\sc HyperNet} \cite{KongT-conf2016a}.  Here, we use the average f-measure (AFM), mean average precision (MAP), mean average recall (MAR), and average intersection of union (AIOU) to quantify performance.  Higher values are better. (f) Ablation study statistics across all of the datasets, which highlight the contributions of each network design decision.  The best performance is obtained when using the {\sc LFN} to provide contextual focus within an \`{a}-trous-deconvolution-based {\sc DNSS} and extracting multi-scale features in an {\sc NRMN}.  The remaining results throughout this paper are reported for this configuration.\vspace{-0.4cm}}
    \label{fig:7}
\end{figure*}

\begin{figure*}
   \includegraphics[]{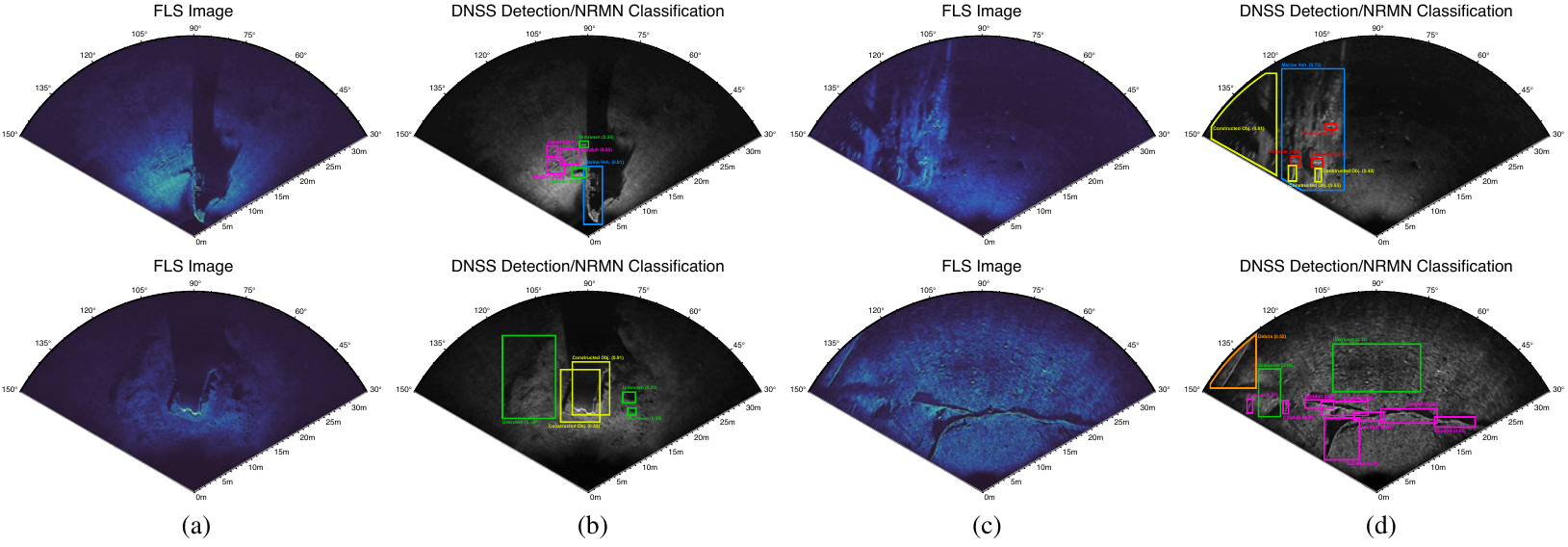}\vspace{-0.05cm}
   \begin{center}\begin{tabular}{l}\hspace{0.3cm}\includegraphics[width=4.6in]{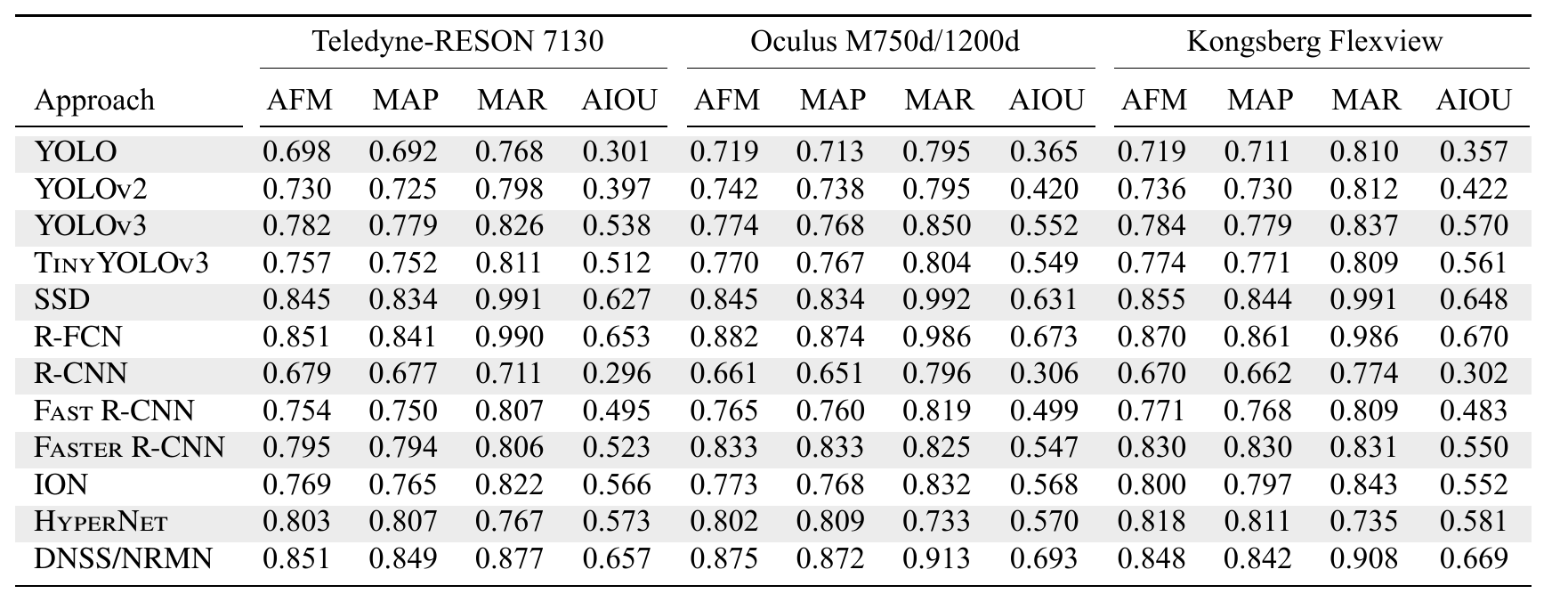} \includegraphics[width=1.1in]{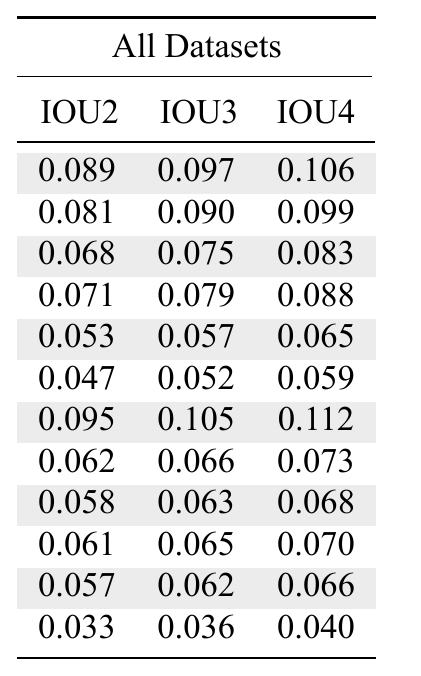}\vspace{-0.175cm}\\
   \hspace{2.5in}{\footnotesize (e)} \hspace{2.55in}{\footnotesize (f)}
   \end{tabular}\end{center}\vspace{-0.2cm}
   \caption[]{\fontdimen2\font=1.55pt\selectfont An overview of the viewpoint insensitivity of the different detectors when trained on one to two passes of large-scale targets and then presented with alternate views of those same targets for additional sensory-platform passes.  The target-platform aspect angle often changes greatly.  Here we consider a ten-shot, multi-way learning strategy.  (a), (c) False-color FLS images of underwater scenes from the test set, which are displayed in a polar-wedge format.  These images have been adaptively despeckled and adaptively contrast equalized.  No temporal aggregation has been performed to enhance shape resolution.  The training set was composed solely of FLS imagery of pass from a vastly different vehicle heading.  (b), (d) The corresponding detection boxes and target confidences, as returned by the {\sc NRMN} for a five-shot, five-way learning strategy.  Each class is denoted using a distinct color.  (e) Table of detection and recognition statistics.  Here, we use average f-measure (AFM), mean average precision (MAP), mean average recall (MAR), and average intersection of union (AIOU) to quantify performance.  Higher values are better.  For all of these statistics, we exclude small-scale targets, since they tend to exhibit only slight variability for different aspects.  (f) Table of detection-box-stability statistics across time.  Here, we report average absolute changes in the AIOU, when they occur inappropriately, across two (IOU2), three (IOU3), and four (IOU4) consecutive frames for correctly localized targets.  Lower values are better, as they indicate that the targets are tracked well and that the bounding box does not change drastically due to local temporal variations in image content. \vspace{-0.4cm}}
   \label{fig:8}
\end{figure*}

\vspace{0.15cm}\noindent {\small{\sf{\textbf{Simulation Results.}}}} Example detection and classification results, along with comprehensive statistical summaries, are provided in \cref{fig:7}.  These results suggest that our framework does well in distinguishing between targets and non-target-like distractors despite the challenging and starkly varying visual characteristics of the FLS imagery.  That is, the {\sc DNSS} fixates well on the former, which permits the specification of tight bounding boxes that overlap well with observed targets; this can be inferred from both the overlaid detection results in \cref{fig:7}(b) and the high average intersection-of-union scores in \cref{fig:7}(e).  Target types of all sizes are adequately isolated, though the {\sc DNSS} performs noticeably better for larger-scale targets, which was also observed in \cref{fig:2}(e).  The {\sc DNSS} is mostly invariant to target orientation, which can be inferred from the average f-measure and precision metrics in \cref{fig:8}(e).  As indicated in \cref{fig:7}(f), the {\sc LFN} aids considerably in maintaining spatio-temporal detection focus for both large and small, static and mobile targets.  Without it, the detection rate drops across all target types.  The lack of temporal regularization conspicuously impacts smaller-scale targets, as they may visually appear as ambient noise in a given FLS image and thus are at risk of being categorized as distractors.

There are some scenes where distractors are incorrectly flagged, by the {\sc DNSS}, as potentially being targets.  In these instances, the {\sc NRMN} often successfully discerns that they do not belonging to one of the established target classes, let alone any of the available sub-classes, and thus are part of some unknown class.  The {\sc NRMN} also accurately labels target-like regions, regardless of the target-vehicle aspect angle.  The memory-based network hence generalizes well from its limited training set size; this claim is corroborated by the average precision and recall scores in \cref{fig:7}(e) and \cref{fig:8}(e).

Our tri-network framework compares favorably to a variety of alternate, deeper networks like {\sc YOLO} \cite{RedmonJ-conf2016a,RedmonJ-conf2017a}, {\sc SSD} \cite{LiuW-conf2016b}, {\sc R-FCN} \cite{DaiJ-coll2016a}, {\sc R-CNN} \cite{GirshickR-conf2014a}, and {\sc Fast R-CNN} \cite{GirshickR-conf2015a,RenS-coll2015a}.  As shown in \cref{fig:7}(e) and \cref{fig:8}(e), when given access to the full training set, which contains several hundred instances of a given class, these networks can recognize targets better than the joint {\sc DNSS}/{\sc NRMN}, albeit not by much.  The detection statistics are, however, sometimes lower for these alternate deep networks.  Single-stage detectors often produce small, superfluous bounding boxes for certain target types instead of larger ones.  This reduces their average precision and intersection of union.  Both the single- and multi-stage networks are sensitive to obvious distractors, such as seafloor morphology, unlike the {\sc NRMN}.  Some networks, like {\sc YOLO} and {\sc R-CNN}, cannot adequately handle small-scale targets, a well-known flaw.  This fact can be gleaned from the statistics for the Teledyne-RESON dataset in \cref{fig:7}(e).  The remaining datasets have mostly large, well-defined targets, in comparison, so the statistics are often better for all of the methods.  From a limited sample set, these alternate networks do not appear to always extract mostly-viewpoint-invariant features, which is corroborated by \cref{fig:8}(e).  These alternate deep networks also do not possess any cross-image, temporal regularizers, like the {\sc LFN}, for propagating region proposals to consecutive images.  The detected regions from these alternate networks can thus fit a target well in one FLS image and then drastically change in the next, due to local spatio-temporal variations in image content, further lowering their precision.  Contextual focus on certain target types can also be lost without such a regularization process.

\vspace{0.15cm}\noindent {\small{\sf{\textbf{Results Discussions.}}}} Our simulation results indicate that {\sc DNSS}s can delineate target-like regions well.  The\\ \noindent {\sc NRMN}s are subsequently able to annotate them with high confidence, using few samples.  In what follows, we outline some of the traits that contributed to the success of our framework.

As was noted in \cite{LinTY-conf2017a}, single-stage networks, such as {\sc YOLO} and {\sc SSD}, sometimes have difficulties delineating and recognizing targets well.  This can be attributed to target-background imbalance during training.  While these networks evaluate a great many candidate locations per FLS image, only a few of those regions contain targets.  This leads to two issues.  The first is that leveraging more samples during training often does little to improve recognition rates.  The imagery contain mostly easily-classifiable, non-target regions that do not contribute to learning.  We see this in our simulations.  The seafloor dominates the FLS imagery, and there is a great amount of textural homogeneity, let alone textural simplicity, in this background class.  Images with targets only comprise a fraction of the overall image stream, even within our curated, temporally-localized datasets.  What targets are present often occupy a small fraction of the entire image.  The second issue is that the prevalence of a great many easy-negative samples can overwhelm training, yielding degenerate models.  We used class-imbalancing heuristics to partly overcome both issues.  However, there is still a potential for these detectors to predominantly focus on so-called easy-learnable and shallow-learnable samples \cite{MangalamK-conf2018a}, like large-scale targets, and hence forgo the more prevalent, small-scale ones.  Robust hard-negative and hard-positive mining \cite{ShrivastavaA-conf2016a} approaches would be more appropriate.  Since we did not need them in our framework, it would not be a fair comparison to make them available to the single-shot networks.

Two-stage frameworks, like ours, {\sc R-FCN}, and {\sc Fast}/{\sc Faster R-CNN}, are less susceptible to these issues.  They cull the number of candidate target locations, in the first stage, removing many background samples.  For {\sc DNSS}s, almost all of the seafloor is filtered, leaving a small region set to be evaluated and converted to bounding boxes.  Learning can mainly occur on the harder examples.  In the second stage, the target and non-target class sample sizes are further harmonized.  In our case, this behavior is realized by our use of a balanced cross-entropy loss to tune the {\sc DNSS} parameters.  This loss downplays the importance of samples with small errors, which are often distributional inliers.  These inliers are large in number, far from the decision boundary, and hence easy to learn.  It instead emphasizes the contributions of samples with large errors, which are often distributional outliers that are few in number and hence hard to characterize with the current filter parameters.  This loss implements an implicit maximum-entropy regularizer that sidesteps making over-confident target location predictions.  The regularizer also exploits unbiased sufficient statistics that can be obtained from the biased samples \cite{DudikM-coll2015a}, thereby enhancing model calibration.  Analogous functionality is not available in any of the comparative deep networks, which partly explains the high detection rates from {\sc DNSS}s.  Moreover, the adaptive error re-weighting realized by our loss function avoids exploding gradients, unlike naive class-weighting schemes.  It thus promotes quick, stable convergence.

These behaviors partly explain the poor performance of the alternate deep networks, especially the single-stage variants.  The sensitivities of the alternate networks to distractors arises from the under-representation of distractors in the training set and their visual similarity with certain sub-classes.  The networks' inability to effectively handle small targets occurs for similar reasons.  As well, it stems from the coarseness of the features in the deeper layers of the networks and the lack of feature sharing from earlier layers.  Other issues are to blame for the poor detection performance.  Networks like {\sc YOLO} have a fixed grid-cell aspect ratio, so they struggle to generalize to targets in unfamiliar configurations.  This explains the poor viewpoint invariance and why multi-aspect detectors, like {\sc SSD} and {\sc YOLOv3}, did better.  {\sc DNSS}s do not suffer from this issue, since they do not rely on grid-cell-based anchor boxes.  The saliency maps are also not downsampled much, so a rich feature set remains for constructing bounding boxes.  These features retain positional information about the targets due to our use of average pooling.  For the alternate networks, the feature sets are derived via max pooling and thus the detections are susceptible to speckle noise.  The derived bounding boxes can hence change greatly across consecutive FLS images.  A lack of temporal regularization for constraining the cross-image deformation of bounding boxes compounds this issue.

These alternate deep networks generate class-agnostic region proposals that are evaluated to regress bounding boxes and produce a classification response.  If a target is not characterized well by invariant features, and hence recognized by the classifier portion of the network, then there is a chance that the corresponding region proposal will be ignored.  Recognition rates are hence adversely impacted, especially for targets whose appearances can change drastically depending on aspect angle.  Saliency detection, in contrast, will retain target-like regions more frequently, regardless of the target type and appearance, provided that the regions are distinct in relation to the local visual content.  If the target-like regions cannot be recognized as belonging to an existing class, then the open-set nature of our {\sc NRMN}s will label them as such.  This functionality is invaluable when dealing with novel targets, particularly in sample-constrained settings.  It is not always straightforward to extend these alternate deep networks to operate reliably in an open-set fashion.  They are thus largely ineffective for zero-shot learning, unlike saliency-based detection.  These alternate networks also cannot be easily scaled to the low-shot case without, for instance, extracting meta-features or adopting feature re-weighting.

On the recognition side, there are a few reasons why our framework does well.  By decoupling the processes of detection and recognition, and considering a relatively shallow architecture for the latter, we inherently preserve ample information about targets.  Such rich features can be further non-parametrically transformed, for a query sample, into a classification response.  This partly justifies why the {\sc NRMN}s can recognize targets, even from novel aspect angles, when trained on so few domain-specific samples.  Very deep detector-recognition networks, in contrast, have the chance of removing discriminative details, in later layers, as the feature set becomes more coarse.  What few features remain may not be descriptive enough to generalize effectively to novel views.  Moreover, for these alternate networks, the uncovered features must be suitable for both detection and classification, whereas, in our case, the {\sc NRMN} features are solely focused on classification.  This promotes target sensitivity and non-target specificity.  The use of external memories, with an appropriate deep attention mechanism, also helps.  Knowledge about the classes is predominantly encoded within the memory entries, not entirely in the convolutional kernels.  The expressive power of the {\sc NRMN}s can be enhanced by simply storing more target entries.  The feature richness and semantic content is not compromised when doing so.  This enables the {\sc NRMN}s to address novel situations.  Conventional network designs, in contrast,  must employ either wider or deeper architectures to achieve the same effect, which, in the latter case, can impede classification due to rapid layer-wise information loss.

Other design decisions contributed to the favorable performance of our framework, as we showed in the ablation study.  The use of \`{a} trous convolution and deconvolution reduced the number of parameters needed to implement robust saliency-based segmentation within the {\sc DNSS}.  The {\sc DNSS} thus could learn effectively from fewer samples than when conventional convolution and unpooling-deconvolution layers were incorporated into the network.  This permitted defining fewer, larger target-region bounding boxes, which improved detection rates.  The integration of {\sc LFN}-based flow fields within the {\sc DNSS} provided pseudo-temporal regularization for detection and recognition, which aided in degraded situations.  It also reduced the number of training samples needed to reach a given performance threshold, since the flow fields could often correct poor target-region initializations, thereby guiding parameter selection during learning.  For the {\sc NRMN}s, aggregating features using a write controller, versus simply storing each training sample in a new memory location, helped with generalization and, surprisingly, viewpoint insensitivity.  Extracting multi-scale features in the {\sc NRMN}s facilitated the recognition of variably-sized targets.  It is likely that more scales and further context modeling are needed, though, to handle the small targets found in the Teledyne-RESON dataset.  As well, more temporal details should be incorporated into the network inference to enhance performance.

Perhaps the greatest driver of performance is the external, recurrent memory.  Conventional deep networks can generalize only through their convolutional kernels.  Wider and deeper convolutional layers are typically needed to build progressively more discriminative characterizations of targets, that capture changes in appearances and multiple viewpoint configurations, along with handling increasing numbers of target types.  Each additional layer can be viewed as part of an internal pseudo-memory for distinguishing between targets.  Many samples are needed to meaningfully stabilize the filter weights, however.  We have demonstrated here that deep hierarchies are not necessarily needed if target representations can be explicitly stored and recalled in external memory and subsequently transformed.  Such representations can be comparatively shallow, since the generalization potential of networks like {\sc NRMN}s do not lie in the complexity of the feature representation.  Rather, it is a function of the number of losslessly retained target depictions.  Every memory entry adds to the complexity and expressive power of the network.  Memories can help generalize to new situations, often without the need to modify the remainder of the {\sc NRMN}.  Moreover, they avoid the overhead associated with explicitly constructing deeper networks, which naturally facilitates low-sample learning.  The incorporation of a write controller for modifying existing memory entries serves to increase target sensitivity and simultaneously compress the features for related targets.  Doing so keeps the overall memory size low while enhancing network expressiveness and preventing overfitting.

\subsection*{\small{\sf{\textbf{4.2.2.$\;\;\;$Comparative Low-Sample Generalization Performance}}}}

We now compare the performance of {\sc DNSS}-guided {\sc NRMN}s with established zero- and low-shot learners from the literature.  No alternate low-sample frameworks are currently available for any imaging-sonar modality.

\vspace{0.15cm}\noindent {\small{\sf{\textbf{Simulation Protocols.}}}} For each of the zero- and low-shot learners, we rely on their standard backbone architectures for feature extraction.  For the zero-shot case, we pre-train them on aPY, SUN, AWA, and ImageNet.  The first three datasets have thorough text-based attributes, and we provided similarly dense class descriptions for the latter using a skip-gram language model trained using {\sc Word2Vec} \cite{MikolovT-coll2013a} on the Wikipedia corpus \cite{ChangpinyoS-conf2016a}.  We then fit the networks to the FLS datasets.  We specify forty text attributes to describe the major classes and sub-classes for the sonar imagery.  For the low-shot case, we pre-train on miniImageNet before fitting to the FLS datasets.  Unless otherwise stated in the references, we employ ADAM-based back-propagation gradient descent with mini batches for training.  We use hyperparameter values listed in the corresponding references, where available; there are too many to succinctly list.

Zero-shot learning assumes disjoint training and test classes.  Traditional testing setups did not make this distinction, though, when pre-training and then fitting the networks.  Some of the sub-classes present in the FLS imagery overlap with those in the above pre-training datasets, even though their visual depictions are distinct.  Despite this, the accuracy for the overlapping classes can sometimes be artificially higher than the others, skewing the statistics.  When reporting results, we consider this so-called standard-split case along with the more general proposed-split configuration where no class overlap is imposed.

We consider the same pre-training dataset and fitting split as in the previous simulation study.  The exception is the one-shot case; for that, we the choose the sample that is most similar to all others in a given class.  We use the same termination criteria unless certain methodologies explicitly considered alternate strategies.  We report our statistics on the combined test and validation sets.  All statistics are averaged over across five Monte Carlo runs.  We exclude runs where learning stalled and hence terminated early.

\begin{figure*}
   \includegraphics[]{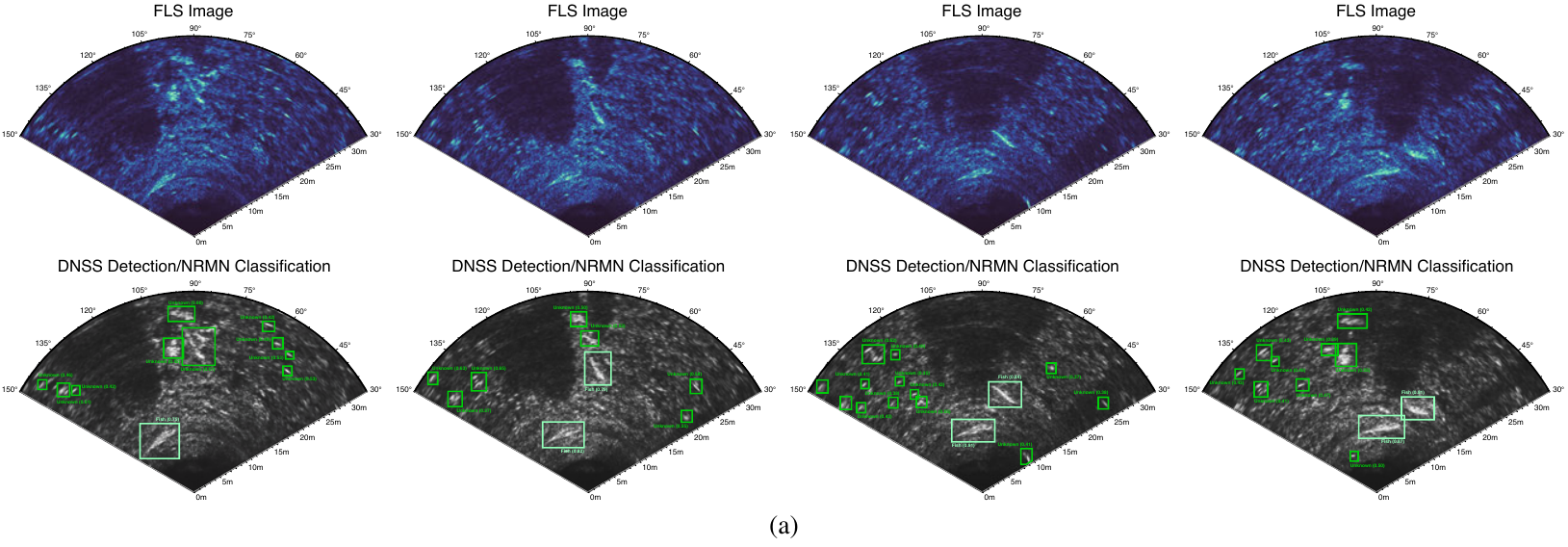}\vspace{-0.475cm}
   \begin{center}\begin{tabular}{l}\hspace{-0.3cm}\includegraphics[width=5.395in]{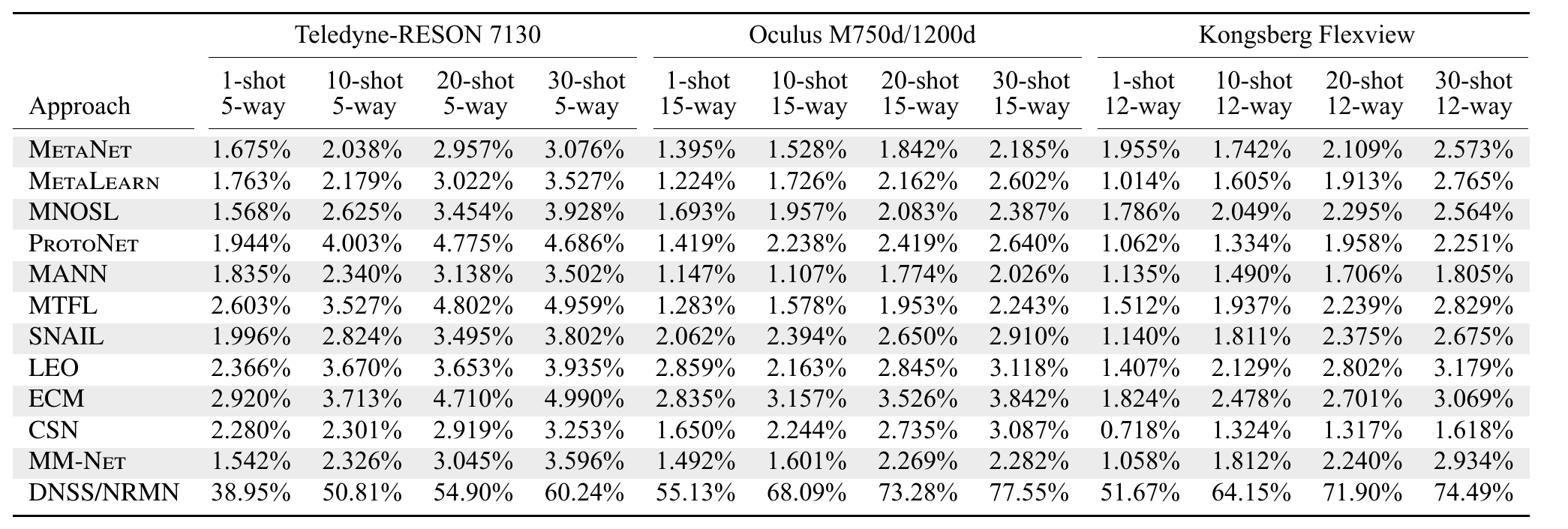}\includegraphics[width=1.34in]{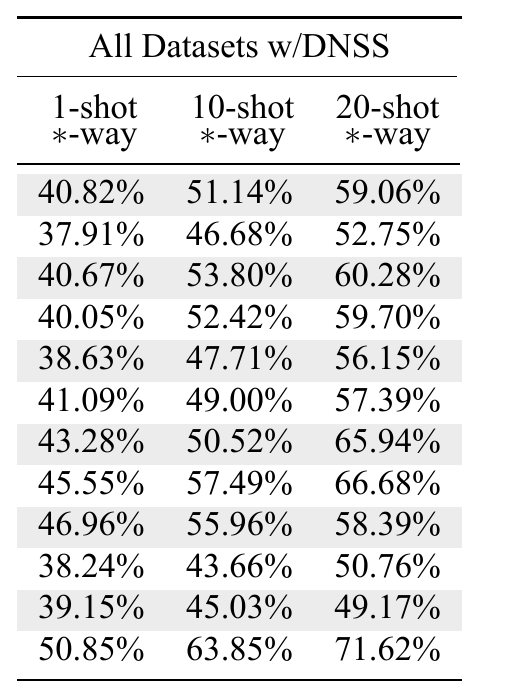}\vspace{-0.175cm}\\
   \hspace{2.65in}{\footnotesize (b)}\hspace{3.05in}{\footnotesize (c)}
   \end{tabular}\vspace{-0.3cm}
   \end{center}
   \caption[]{\fontdimen2\font=1.55pt\selectfont A comparison of {\sc DNSS}-guided {\sc NRMN}s with other low-shot learners.  (a) False-color FLS images of underwater scenes from the test set, which are displayed in a polar-wedge format.  Here, two fish swim near a rocky bottom type.  These images have been adaptively despeckled and adaptively contrast equalized.  No temporal aggregation has been performed to enhance shape resolution.  The training set was composed solely of FLS imagery of pass from a vastly different vehicle heading.  Beneath each image we display the detection boxes, from {\sc DNSS}, and target confidences, as returned by the {\sc NRMN} for a twenty-shot, multi-way learning strategy.  Each class is denoted using a distinct color.  (b) Table of recognition statistics for different numbers of training samples per class, which are referred to as shots.  Higher values are better.  Here we consider {\sc MetaNet} \cite{MunkhdalaiT-conf2017a}, {\sc MetaLearn} \cite{RaviS-conf2017a}, {\sc MNOSL} \cite{VinyalsO-coll2016a}, {\sc ProtoNet} \cite{SnellJ-coll2017a}, {\sc MANN} \cite{SantoroA-conf2016a}, {\sc MTFL} \cite{SunQ-conf2019a}, {\sc SNAIL} \cite{MishkinD-conf2018a}, {\sc LEO} \cite{RusuAA-conf2019a}, {\sc ECM} \cite{RavichandranA-conf2019a}, {\sc CSN} \cite{KochG-conf2015a}, and {\sc MM-Net} \cite{KaiserL-conf2017a}.  None of the alternate low-shot learners can handle multi-target classification, so we report whenever they correctly label images with just a single target.  This does not occur often.  (c) Detection statistics, over all datasets, when giving each approach access to the detection boxes obtained from the {\sc DNSS} saliency maps.  The alternate methods become more competitive in this case.\vspace{-0.4cm}}
   \label{fig:10}
\end{figure*}

\vspace{0.15cm}\noindent {\small{\sf{\textbf{Simulation Results.}}}} Example low-shot detection and classification results, along with comprehensive statistical summaries, are provided in \cref{fig:10}.  When compared with the results in the previous section, those in \cref{fig:10}(b) highlight that, when fit to just a single sample per class, our framework is competitive with {\sc YOLO}, {\sc YOLOv2}, {\sc R-CNN}, and {\sc Fast R-CNN}.  As the number of shots increases, the recognition rate improves immensely, even when distinguishing between many classes.  It is thus competitive with {\sc SSD} and {\sc R-FCN}, the best two networks that we considered.  This is despite that our framework has a much smaller feature-extraction backbone.  Nevertheless, the {\sc NRMN}'s backbone is sufficiently complex enough that mostly-aspect-angle-insensitive features can derived and transformed.  It likewise enables the labeling of differently-sized targets.  The {\sc NRMN} handles moving targets well too, as we show for the example scene in \cref{fig:10}(a).  Such targets can be adequately discerned even when the sensing platform shifts position and rotates.  Contextual focus is only lost as the target goes out of view.  Large-displacement movements that occur rapidly prove too challenging, though, for our current framework.  They confound the other deep detectors too.

Our framework substantially outperforms many other deep, low-shot learners, as highlighted in \cref{fig:10}(b).  These approaches can only recognize a single class in a given image, by default.  It is rare that only a single instance from a single class is present in an FLS image.  When given access to the {\sc DNSS}-derived detection boxes, their performance is better, as shown in \cref{fig:10}(c), though it often trails behind that of the {\sc NRMN}.  This is because these alternate networks cannot generalize well to targets whose appearance and size varies across time.  They also cannot extract sufficient discriminative details from smaller-sized targets.  They are thus only competitive with deep detectors like {\sc YOLO} and {\sc R-CNN}, unlike our framework.  The exceptions are {\sc LEO} and {\sc ECM}.  None of these networks support open-set classification, so they make many mistakes when leveraging the supplied region proposals.  Our {\sc NRMN}s, in contrast, simply label many distractors as unknown entities, thereby keeping recognition rates high.

\begin{figure*}[t!]
   \includegraphics[]{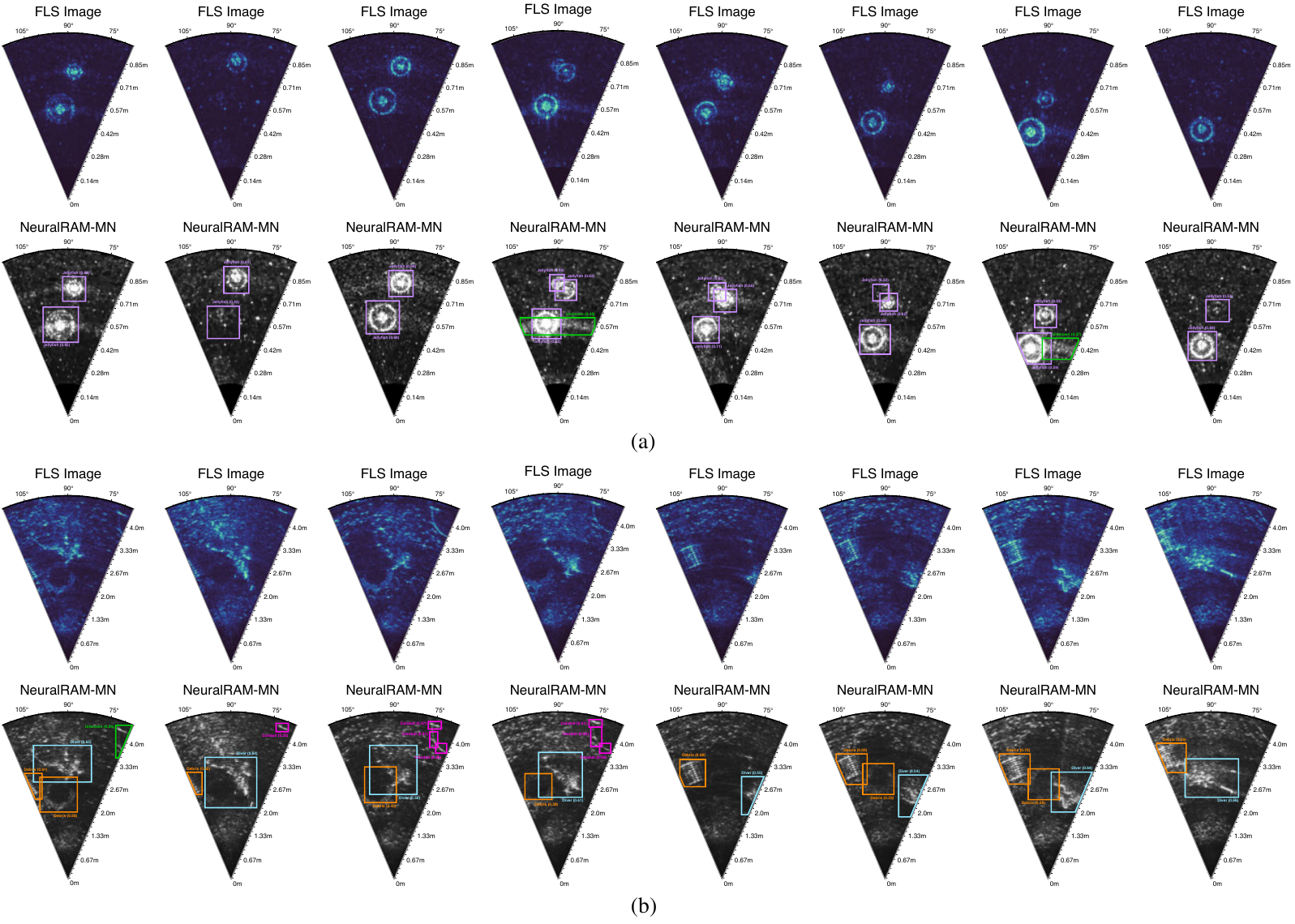}\vspace{-0.05cm}
   \begin{center}\begin{tabular}{c}\hspace{-0.3cm}\includegraphics[width=6.25in]{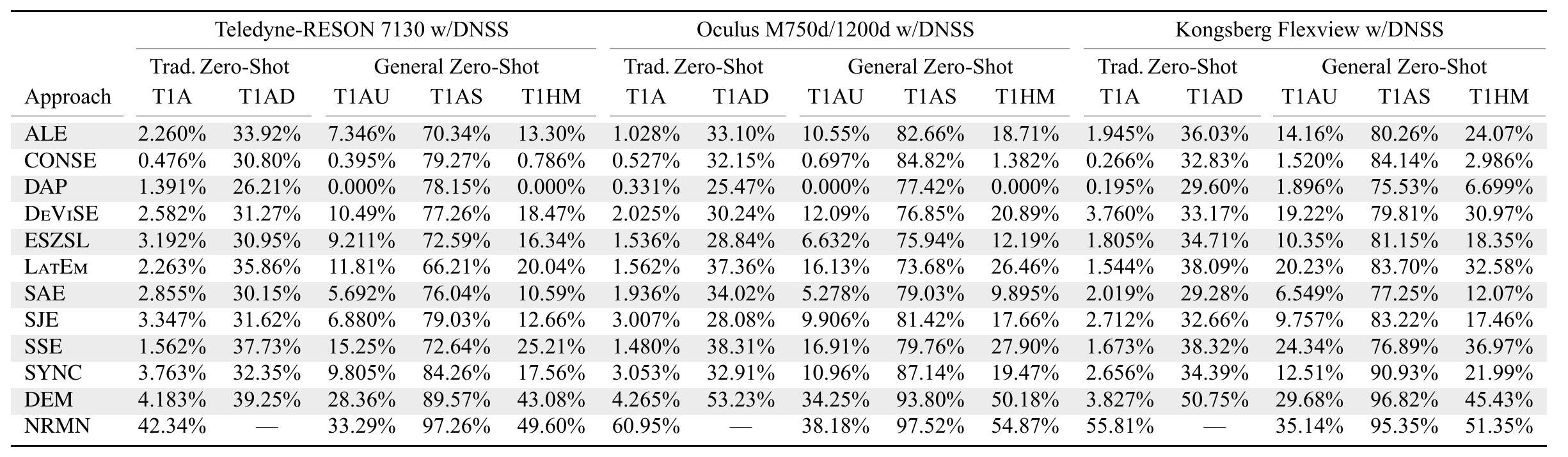}\vspace{-0.175cm}\\
   \hspace{-0.45cm}{\footnotesize (c)}
   \end{tabular}\vspace{-0.3cm}
   \end{center}
   \caption[]{\fontdimen2\font=1.55pt\selectfont An overview of zero-shot learning with a modified version of our {\sc NRMN}s that relates semantic content with existing representation embeddings.  (a) An example of a never-before-seen class, a jellyfish, that is accurately detected, tracked, and recognized.  (b) Another instance of a never-before-seen class, a human diver, that is again accurately detected, tracked, and recognized.  For both (a) and (b), the top rows contain false-color FLS images of underwater scenes from the test set, which are in a polar-wedge format.  We have cropped the horizontal aperture of the FLS images, after processing, to compactly display more imagery; the full aperture range is processed by our {\sc NRMN}s, though.  The bottom rows contain the class-color-coded target detection bounding boxes and classifier confidences. (c) Table of recognition statistics for a variety of alternate zero-shot learners trained on each dataset: ALE \cite{AkataZ-conf2013a}, CONSE \cite{NorouziM-conf2014a}, DAP \cite{LampertCH-jour2014a}, {\sc DeViSE} \cite{FromeA-coll2013a}, ESZSL \cite{RomeraParedesB-conf2015a}, LATEM \cite{XianY-conf2016a}, SAE \cite{KodirovE-conf2017a}, SJE \cite{AkataZ-conf2015a}, SSE \cite{ZhangZ-conf2015a}, SYNC \cite{ChangpinyoS-conf2016a}, and DEM \cite{ZhangL-conf2017a}.  Here we report the top-one-percent accuracy on both the full image (T1A) and on the detection boxes returned by the {\sc DNSS} (T1AD).  None of these zero-shot learners can handle multiple targets in a single image, unlike our framework, so the T1A is naturally low.  We also report the T1A on the unseen (T1AU) and seen (T1AS) samples, which is the general zero-shot-learning setting proposed in \cite{XianY-conf2017a}.  We also report the harmonic mean (T1HM) of the T1AU and T1AS statistics, which is a good measure overall performance.  Higher values are better.  For the T1AU, T1AS, and T1HM, we trained and applied the alternate methods to the target-like regions from the {\sc DNSS}. \vspace{-0.4cm}}
   \label{fig:11}
\end{figure*}

Additional low-shot results for our framework, in the form of zero-shot generalizations, are given in \cref{fig:11}.  These results illustrate the promise of the {\sc DNSS} for broad target identification in sample-deficient settings.  Despite not encountering these target classes before, our network could isolate instances from them well and track them over time.  None of the alternate deep detectors that we consider offer this functionality, as they only make closed-set decisions.  These results also indicate that, when properly pre-trained, the {\sc NRMN}s can recognize novel target classes by relating text descriptions of them to both visual and textual depictions of previously encountered targets.  Text is, naturally, no substitute for imagery, so the performance is lower than in the one- and multi-shot cases.  However, it is easier to specify, so it offers the potential to broaden models to new applications, where potentially no data is initially available, and later fit them as new labeled samples are obtained.  As with the low-shot cases, assigned target labels are mostly stable, across consecutive FLS images, but only if the target appearance does change drastically.  If the visual characteristics could not be related well to the text descriptions, then the {\sc NRMN} would assign those regions to the unknown class. 

None of the zero-shot learners to which we compare can accommodate multiple targets in a single sample.  As shown in \cref{fig:11}(c), their top-one accuracy is quite poor.  It is only when given access to the {\sc DNSS} detected regions that they begin to be somewhat effective for recognition.  However, their performance is quite poor, as they make unrealistic assumptions about how to combine disparate content to arrive at a class label.  Much of their accuracy, for the novel classes in the FLS test-set imagery, appears to be derived from haphazardly guessing the correct label.

\vspace{0.15cm}\noindent {\small{\sf{\textbf{Results Discussions.}}}} Our simulation results indicate that {\sc NRMN}s are effective at generalizing from few and even no samples.  In what follows, we describe why some existing learning schemes failed to behave as well.

Memory-based matching networks offer several benefits for reducing network complexity while promoting network discriminability.  Some existing networks, though, have issues that impede this objective.  Networks like {\sc MNOSL} implement {\sc LSTM}-based memories that naively store feature representations.  They directly store the sample features and do not possess an efficient writing mechanism to combine related entries and enhance their expressive capability.  They also retain infrequently and never-queried entries, even if they do not contribute to recognition performance.  This can prevent effective generalization in fixed-size memories.  It poses significant difficulties for continuous, lifelong learning.  As well, the read operations for {\sc MNOSL} and {\sc MANN} are based on soft attention and do not scale well in terms of either the number of entries used or the dimensionality of stored features.  Moreover, {\sc MNOSL}, {\sc MANN}, and {\sc ProtoNet} generalize only via convex combinations of stored values.  This burdens the controller with estimating useful key and value representations.  It also presupposes that the target attributes are mostly linearly separable, which is not explicitly guaranteed in the convolutional feature extraction portion of these networks.  It is also not directly imposed by their choice of loss function.  The correct classes may not be predicted as a consequence.  We avoid this issue here by learning a deep similarity measure, which adaptively and non-linearly transforms the memory entries to uncover an appropriate label.

We heavily discourage the use of {\sc LSTM}-based memories for many low-shot-learning settings.  Their memory size is fixed and cannot be extended without significant network retraining.  They cannot handle complex spatio-temporal problems well, either, which complicates higher-level target analysis like target re-identification.  This is because {\sc LSTM}s have temporally linear hidden-state dependencies.  {\sc NeuralRAM} cells do not have these disadvantages.  As well, many of the  {\sc NeuralRAM}s' training stability issues have been fixed.

{\sc SNAIL} \cite{MishkinD-conf2018a} is an oddity out of the many approaches we consider.  It is a convolutional network that attempts to behave, in some circumstances, like a recurrent one.  It is composed of dilated, causal convolutional layers that are coupled with a soft attention mechanism.  Both components allow {\sc SNAIL} to generalize over fixed-size temporal contexts, sometimes even more effectively than conventional recurrent networks.  {\sc SNAIL} cannot compete with external memories, though.  Its ability to remember far into the past is tied to an exponentially-increasing dilation rate.  {\sc SNAIL} networks thus often only have coarse access, if any at all, to such entries.  Moreover, the number of layers needed to recall representations of past samples scales logarithmically with the sample amount.  Although soft attention can help within pinpointing a class-based representation, even from a potentially infinite context, its utility is somewhat limited in {\sc SNAIL}.  This is because there are implicit constraints on how much the underlying convolutional network can retain and recall.  It is doubtful that the retained content is losslessly encoded, either, which introduces the chance for classification mistakes as more samples are presented during learning.  Our framework does not have these issues.  {\sc NeuralRAM}s can be resized to store any number of entries losslessly.  Once trained well, our deep, non-parametric similarity measure is agnostic to the entries.  It simply discerns if they are well related to the query samples.  This measure can handle increases and decreases in memory content, including the incorporation of new target classes, without retraining.  This does not appear to be true for {\sc SNAIL}'s attention mechanism.

There are other options for enabling low-sample generalization, many of which are either model-based, optimization-based, or hybrids.  {\sc MetaNet}s, for instance, define fast and slow weight branches within each network layer and train a meta-learner that generates fast weights for one-shot adaptation of a classification network.  Their use of an {\sc LSTM} complicates learning, though.  It also impedes continuous learning, since more {\sc LSTM} cells must be added to learn more things, which requires substantial re-training.  As well, if the derived meta information is not independent from the task space, then {\sc MetaNet}s may exhibit poor generalization to new problems.  This appeared to happen here.  Another optimization-based approach is {\sc MAML} \cite{FinnC-conf2017a}.  {\sc MAML} discovers a network parameter initialization that can be rapidly adapted to several related tasks by executing a few steps of gradient descent.  Probabilistic extensions to {\sc MAML} have recently been proposed.  These versions are trained via a variational approximation that uses simple posteriors.  It is not immediately clear how to extend these posteriors to more complex distributions, with a more diverse set of tasks, meaning that their utility for FLS imagery is low.  Additionally, most instances of {\sc MAML}, along with other optimization-based meta learners, are highly prone to overfitting.  This is because they rely on only a few samples to compute gradients in high-dimensional parameter spaces, which makes generalization difficult, especially under the constraint of a shared starting point for task-specific adaptation.  We avoid overfitting, to some extent, in our framework, which is due to our use of a memory-modifying write controller.  This controller constructs weighted-average-like depictions of targets.  It also selectively forgets infrequently-used entries over long learning periods.

Recognizing this overfitting issue, some approaches train a deep input representation, or concept space, and use it as input to a meta-learning network.  They conduct gradient-based adaptation directly in the parameter space, which is still comparatively high-dimensional.  Finding good parameters in such a space can be difficult when considering few samples.  Learning in a low-dimensional latent space and performing optimization-based adaptation in it, not a parameter space, leads to superior generalization.  Few samples are needed to handle novel target appearances and viewpoints.  This is partly why an approach like {\sc LEO} \cite{RusuAA-conf2019a} has higher recognition rates.

{\sc LEO} is one of the few methods that is competitive against {\sc NRMN}s despite not extracting multi-scale features.  Another is {\sc ECM} \cite{RavichandranA-conf2019a}.  {\sc LEO} possesses two key advantages over other optimization meta-leaners.  First, it conditions the initial parameters for a new task on the training samples.  This leads to a task-specific starting point for adaptation.  The parameters are often much closer to good local optima than when considering either random initializations or initializations in high-dimensional spaces.  {\sc LEO} additionally incorporates a relational sub-network, much like we do in the {\sc NRMN}.  This sub-network maps the few-shot samples into a latent vector space.  Doing so allows {\sc LEO} to consider context when initializing parameters.  It essentially is cognizant that decision boundaries required for fine-grained distinctions between similar classes may be different from those for broader classification.  Secondly, as we mentioned, {\sc LEO} optimizes in a lower-dimensional latent space, which adapts the model behavior more effectively than when considering the parameter space.  By forcing this adaptation to be stochastic, the ambiguities present in the few-shot samples can be captured.  One shortcoming of this latent space, at least as it is implemented in {\sc LEO} and other networks, is that its dimensionality is fixed across all tasks.  Varying the degrees of freedom depending on the task, or even the class, would be a more appropriate choice and likely would improve performance.  Another shortcoming is that a task-specific subspace metric is not learned.  We do both in the {\sc NRMNs}.

{\sc ECM} does well since it possesses some similar functionality as {\sc NRMN}s.  {\sc ECM} learns a metric, albeit a non-deep one.  This allows it to account for novel target types without crowing the class representation space.  While some researchers believe that point-estimated metrics are highly sensitive to noise \cite{ZhangJ-conf2019a}, especially in low-sample settings, we did not see this behavior materialize.  {\sc ECM} also uncovers class identities, which map the features from the same class to some representative in a potentially non-linear fashion.  The memory write controller in the {\sc NRMN}s uncover analogous class-based prototypes.  The distinction is that {\sc NRMN}s can define multiple prototypes.  Doing so enhances nearest-neighbor-based searches in the later stages of the network.  {\sc ECM}, as it was originally proposed, only constructs a single prototype per class, which may not be sufficient for complicated target types.  Low-shot learning should rely on multiple prototypes to increase the potential complexity of the decision boundaries.

Only a handful of low-shot learners can extend immediately to the zero-shot case.  Many of the optimization-based schemes are too brittle to find good parameters that work for never-before-seen classes.

Some of the zero-shot learners that we consider attempt to learn a mapping between the image feature space and a semantic space.  These features are typically derived from some pre-trained convolutional backbone network that is rather deep.  {\sc ALE} \cite{AkataZ-conf2013a}, for instance, learns a bilinear compatibility function, via a weighted, approximate ranking loss, to associate visual and semantic information with target classes.  {\sc DeViSE} \cite{FromeA-coll2013a} also uncovers a linear mapping in a similar, albeit non-weighted, manner as {\sc ALE}.  {\sc SJE} \cite{AkataZ-conf2015a} does too, but it only considers the top-ranked class, not some combination of compatibility functions from each class.  This makes it more prone to classification errors than either {\sc ALE} or {\sc DeViSE}, which is corroborated by our simulation results.  {\sc SAE} \cite{KodirovE-conf2017a} relies on a linear semantic-autoencoder loss to constrain the projection to reconstruct the original image embedding.  The final approach that we consider in this category, {\sc ESZSL} \cite{RomeraParedesB-conf2015a}, applies a square loss to the ranking loss and adds regularization terms.   Owing to the linear nature of these approaches, their performance lags compared to the non-linear, non-parametric similarity measure implemented by {\sc NRMN}s.  As well, constructing linear embeddings is a poor design decision.  While these approaches operate on deep features, there are no feature-separability guarantees for many conventional learning losses.  Attempting to learn a linear classifier in a linearly-constructed latent space that actually performs well is fraught with difficulties.  One of the few practical advantages of linearity is that it preempts overfitting, to some extent, in low-sample settings.  It is more prudent to non-linearly embed the visual and textual content, though, as we do with the {\sc NRMN}.  Non-linearity helps emphasize those feature components that contribute the most to recognition.  The potential for overfitting is outweighed by the decision-making expressivity of the network.

Certain hybrid learners exist that express images and semantic class embeddings as a mixture of seen class proportions.  {\sc SYNC}, constructs classifiers for the novel classes via linear combinations of base classifiers.  {\sc CONSE} and {\sc DAP} rely on similar processes, although they are probabilistic.  This, as we noted, has difficulties modeling complex relationships between the multi-modal content.  Recognition thus rates suffer.  It would be far more sensible to kernelize any of these methods than simply creating permutations of, what amounts to, the same linear framework.  Recognizing the drawbacks of this linear assumption, SSE \cite{ZhangZ-conf2015a} learns a non-linear, sparse-coding-inspired mapping for the semantic embedding and a class-dependent transformation for the visual embedding.  It then uses the mixture of seen class proportions as the common space assigns images to classes based on their mixture patterns.  This method may suffer from data bias, though, leading to unreliable predictions.  It is also sensitive to noisy side information.  As well, it, and all of the other zero-shot learners to which we compare, can only operate on single-word attributes.  Our use of convolutional text embedding additionally allows the {\sc NRMN}s to handle both short and lengthy class descriptions with ease.  Another approach, {\sc LatEm}, extends the {\sc SJE} model and learns a piece-wise linear mapping for the latent variables.  Piece-wise linearity cannot compete with the non-linear nature of the {\sc NRMN}s, though.  This non-linearity arises in the memory-query comparison.  It also occurs in the semantic embedding, as we apply a pre-trained, non-parametric convolutional network to the text descriptions.

None of these zero-shot learners possess memories.  Accounting for new classes thus requires re-optimizing their corresponding loss functions.  For {\sc NRMN}s, new classes can be included by simply embedding the visual and textual content and storing the representations, along with the class label, in the {\sc NeuralRAM} cell.  Unless a great many novel categories are included, an {\sc NRMN} can generalize without any parameter updates, provided that it was pre-trained well.  Re-training an {\sc NRMN} may improve performance, though.

\subsection*{\small{\sf{\textbf{5.$\;\;\;$Conclusions}}}}

We have proposed a novel deep-network framework for zero- and low-shot, multi-target detection and recognition in FLS imagery.  Our framework represents the first instance of low-sample learning for any imaging sonar modality.

There were several crucial insights which enabled this work.  One is that low-sample generalization is feasible for complicated, noisy image-based modalities if distractors are adequately removed.  Another is that low-sample learning and generalization is much easier if the network is both designed around and trained to conduct low-sample learning.  Non-parametric structures, like memory, also facilitate network adaptation to new cases without the training overhead associated with deeper networks.

Based on these observations, we developed a real-time, tri-network framework.  The first network performs saliency-based localization of target-like regions.  Bounding boxes can then be derived from the produced saliency maps to delineate potential targets.  Each of the candidate targets is then cascaded into a memory-based matching network to produce an open-set classification response.  The third network temporally regularizes the saliency maps.  It helps provide consistent detections and labels across consecutive FLS images in a stream. 

We validated our framework on a variety of real-world FLS datasets.  We showed that, with only a few samples, our framework operated on par with deeper networks that could seemingly extract more semantic content from greater amounts of samples.  In some cases, our framework did better.  This performance discrepancy partially arose due to decoupling the processes of detection and recognition, which permitted retaining discriminative target details.  It also arose from our use of saliency-based detection, which avoided class imbalance issues during recognition.  Several other factors contributed too.

We additionally compared our framework with others that can generalize well from few samples.  We demonstrated that ours performs far better, since existing zero- and low-shot approaches cannot presently handle multi-target recognition.  When given access to the same detection bounding boxes that our framework derived, the performance of some of these alternate approaches still lagged.  We attributed this difference to the use of fixed metrics, which pre-suppose near-linear separability of the features.  It also stemmed from the lack of multi-scale features to account for targets of vastly differing sizes.

Our use of external, tabular memories was crucial for keeping network size small while still achieving high target labeling accuracy.  Most deep networks can often only construct class-specific representations in the later stages of the convolutional hierarchy.  Many layers are therefore needed to compose progressively more discriminative target details and thus internally remember traits about them.  External memories shortcut this process to some extent.  They store and recall class-based representations explicitly, not implicitly in the convolutional kernels, the latter of which is a rather poor way of retaining class instances well.  Memories crucially avoid the overhead of learning many filter parameters to store target details.  They are hence very sample efficient.

A current limitation of our tri-network framework is that it can only handle either mobile targets for a mostly stationary sensor platform and mostly stationary targets for a mobile sensor platform.  It can struggle with fast-moving mobile targets for a mobile sensor platforms.  This stems from our choice of convolutional architecture.  That is, we emphasize relatively small-scale networks that extract spatial features, not spatio-temporal features, to detect and recognize targets.  We then leverage some temporal content, towards the end of the processing pipeline where a target call is about to be made, to modify both the bounding boxes and associated target prediction labels.  This fragmented approach conveniently permits pre-training each of the networks separately, reducing the amount of domain-specific imagery needed to obtain good performance in most cases.  It, however, comes at the expense of not exploiting all of the temporal content present in the FLS image streams.  Our future endeavors will therefore emphasize the use of three-dimensional convolutional kernels for maintaining appearance representation quality and modeling long-range temporal dependencies.  The remaining challenge will then be to train these parameter-intensive networks in a way that still allows for low-sample generalization; this is not a trivial task.

Another limitation is that our framework currently only operates on FLS imagery from a single frequency band.  Lower frequency bands have a high beamwidth, and hence a large sampling volume, but exhibit qualitatively poor shape resolution and are often corrupted with high amounts of omni-directional, ambient noise.  They are, however, modestly insensitive to sensory platform motions, as ping-to-ping overlaps in the ensonified volume will occur during pitch and roll motions.  Higher bands are the reverse of this.  Modifying the networks to account for multiple frequency bands is trivial.  Doing this effectively, however, will require significant investigation.  One research avenue will entail specifying a deep network that extrapolates from low to high acoustic frequencies.  It will provide pseudo-high-frequency coverage, thereby increasing the extent of the high-frequency-band FLS imagery in a contextually sensitive manner.  Both the low-frequency and this pseudo-high-frequency FLS imagery can then be processed by a multi-band convolutional network.  The combination of multiple bands should moderately enhance detection and recognition performance for a variety of situations, especially for targets located near the bottom dead zone.   We will additionally incorporate FLS-derived bathymetry to further improve target localization and seafloor separability and hence recognition rates.

A final limitation of our framework is that it learns point-based estimates of class representations.  It fails to codify higher-order statistics, for the multi-shot cases, that may lead to a more complete class depiction.  A more elegant approach would be to infer distributions from the samples.  This could be done via non-parametric kernels so that the convergence to the distributional estimate is independent of the sample dimensionality, a crucial property in sample-constrained settings when processing complex visual content.  The similarity comparison could likewise be conducted within this kernel-based paradigm, thereby simplifying aspects of our framework.

\setstretch{0.95}\fontsize{9.75}{10}\selectfont
\putbib
\end{bibunit}

\end{document}